\documentclass[sigplan,review=false]{acmart}
\renewcommand\footnotetextcopyrightpermission[1]{} 
\usepackage{tikz}
\usepackage{amsmath}

\usepackage{filecontents}
\usepackage{enumitem}
\usepackage{pbox}
\usepackage{graphicx}
\usepackage{subcaption}
\usepackage{appendix}
\usepackage{amsthm}
\usepackage{float}
\usepackage{multirow}
\usepackage[linesnumbered,ruled]{algorithm2e}

\newtheorem*{problem}{Problem}
\newtheorem{lemm}{Lemma}
\newtheorem{theore}{Theorem}
\newcommand{\name}{ByteComp}

\begin{document}
\title{{\name}: Revisiting Gradient Compression in Distributed Training}


\author{Zhuang Wang}
\email{zw50@rice.edu}
\affiliation{%
  \institution{Rice University}
  \country{USA}
}

\author{Haibin Lin}
\email{haibin.lin@bytedance.com}
\affiliation{%
  \institution{ByteDance Inc.}
  \country{USA}
}

\author{Yibo Zhu}
\email{zhuyibo@bytedance.com}
\affiliation{%
  \institution{ByteDance Inc.}
  \country{USA}
}

\author{T. S. Eugene Ng}
\email{eugeneng@cs.rice.edu}
\affiliation{%
  \institution{Rice University}
  \country{USA}
}


\begin{abstract}
Gradient compression (GC) is a promising approach to addressing the communication bottleneck in distributed deep learning (DDL).
However, it is challenging to find the optimal compression strategy for applying GC to DDL because of the intricate interactions among tensors.
To fully unleash the benefits of GC, two questions must be addressed:
1) How to express all compression strategies and the corresponding interactions among tensors of any DDL training job?
2) How to quickly select a near-optimal compression strategy?
In this paper, we propose {\name} to answer these questions.
It first designs a decision tree abstraction to express all the compression strategies and develops empirical models to timeline tensor computation, communication, and compression to enable {\name} to derive the intricate interactions among tensors.
It then designs a compression decision algorithm that analyzes tensor interactions to eliminate and prioritize strategies and optimally offloads compression to CPUs.
Experimental evaluations show that {\name} can improve the training throughput over the start-of-the-art compression-enabled system by up to 77\% for representative DDL training jobs.
Moreover, the computational time needed to select the compression strategy is measured in milliseconds, and the selected strategy is only a few percent from optimal.

\end{abstract}

\maketitle
\pagestyle{plain}


\section{Introduction} 

Deep Neural Networks (DNNs) have brought remarkable success to domains such as computer vision~\cite{ResNet-50, rethinkingCV, vgg, LSTM} and natural language processing (NLP)~\cite{bert, kiros2015skip, manning2014stanford, vaswani2017attention}.
Because today's training jobs with a single GPU typically take days and even weeks~\cite{pipedream, cho2019blueconnect}, 
data-parallel distributed deep learning (DDL) has become the norm to accelerate the training with multiple GPUs~\cite{PSosdi2014, byteps, horovod, tensorflow, pytorch_ddp}.

However, there exists an exacerbating tension between computation and communication in DDL.
The recent innovations of hardware accelerators~\cite{phub, nvidia_performance} and domain-specific software optimization~\cite{chen2018tvm, zheng2020ansor, chetlur2014cudnn} have dramatically reduced the computation time of DNN training.
For example, the single-GPU iteration time of ResNet50 has seen a 22$\times$ decrease in the last seven years~\cite{training_time}.
This trend leads to more frequent gradient synchronization in DDL and puts higher pressure on the network.
However, it is difficult for GPU cloud network deployments to match this pace; network bandwidth has grown only by roughly $10\times$ in the same period~\cite{SwitchML, zhou2020beyond, Mellanox_update, OpenAI, nvidia_performance}.

The growing concern of communication bottlenecks in DDL has motivated numerous works, such as priority-based scheduling~\cite{bytescheduler, p3, tictac}, wait-free back-propagation mechanism~\cite{poseidon, pytorch_ddp}, and optimized aggregation algorithms~\cite{byteps, cho2019blueconnect, horovod}. 
However, even with the latest highly-optimized BytePS~\cite{byteps} which incorporates these state-of-the-art approaches, communications for gradient synchronization still account for 42\% and 49\% of the total training time of GPT2~\cite{gpt2} and BERT-base~\cite{bert} with 64 NVIDIA V100 GPUs in 8 machines connected by a 100Gbps Ethernet network.


Gradient compression (GC) algorithms~\cite{stich2018sparsified, aji2017sparse, dgc, sparse2018gradient, wen2017terngrad, 1bit, efsignsgd, QSGD} have a great potential to address the communication bottlenecks in DDL by saving up to 99.9\% of the gradient exchange
while preserving the training accuracy and convergence~\cite{wu2018error, stich2018sparsified, jiang2018linear}.
However, the training speedups of DDL with GC are only modest because of the costly compression operations.
For example, applying GC to the aforementioned GPT2 training only achieves a $1.15\times$ speedup.
This motivates us to revisit GC from the system perspective to fully unleash its benefits for DDL.



Applying GC to a DNN model entails many decisions for each tensor, such as whether to compress, the type of compute resources for compression and the communication schemes for compressed tensors.
DDL typically involves both communications inside a machine and across machines.
Therefore, another decision is whether to apply GC to intra- or inter-machine communication or both.
The compression strategy, i.e., the decisions for all tensors, determines the training throughput of compression-enabled DDL.

Unfortunately, it is very challenging to make these decisions because of the intricate interactions among tensors.
Therefore, the first research question we have to answer to unleash the benefits of GC is how to express all possible compression strategies and the corresponding interactions among tensors for any DDL training job?
Because of the extremely large search space, even if all the strategies and the interactions are available, the time to find the optimal one can be prohibitive.
Hence, the second research question is how to analyze the interactions among tensors to quickly select a near-optimal compression strategy?

In this paper, we propose {\name} to answer these two questions in order to maximize the benefits of GC.
We make the following contributions.

\noindent $\bullet$ 
We develop a decision tree abstraction for the compression strategy and empirical models for the time of tensor computation, communication, and compression to answer the the first question.
The abstraction can express all possible compression options of any tensors regardless of different tensor sizes and GC algorithms.
Based on the abstraction, {\name} can express all compression strategies of any DDL training jobs.
The empirical models enable {\name} to derive the timeline of tensor computation, communication, and compression of all tensors in a DNN model, and thus their intricate interactions with any compression strategy.

\noindent $\bullet$
We propose a compression decision algorithm for quickly selecting a near-optimal compression strategy to answer the second question.
{\name} analyzes the interactions among tensors to eliminate a large number of suboptimal compression strategies.
Based on the analysis, {\name} proposes a prioritization method for applying GC to tensors to maximize the benefits, and considers the overlapping time among tensor computation, communication, and compression to make compression decisions for each tensor.
Because of different performance trade-offs of GPUs and CPUs for GC, {\name} finds a provably optimal solution to offload compression from GPUs to CPUs to minimize the resource contentions with tensor computation.

\noindent $\bullet$ 
We implement a fully featured system for {\name}. 
We implement both GPUs and CPUs compression libraries.
We also implement communication libraries to support different communication schemes in both intra- and inter-machine communications.
Experimental evaluations demonstrate that with 64 GPUs, {\name} can improve the training throughput by up to 269\% compared with BytePS.
It also outperforms the state-of-the-art compression-enabled system (i.e., HiPress~\cite{HiPress}) by up to 77\% across representative DNN training jobs.
Moreover, the computational time needed by {\name} to select the compression strategy is measured in milliseconds, and the performance difference between the selected strategy and the optimal strategy is only a few percent.



\section{Background}
\subsection{Communication in DDL}
\label{sec:DDL_comm}

In data-parallel distributed deep learning (DDL), each GPU has a replica of the DNN model.
The training dataset is divided into multiple partitions and each GPU takes one partition.
Training is performed in multiple iterations.
At the beginning of an iteration, each GPU consumes a mini-batch of training data from its own partition.
It then independently performs \textit{forward propagation} and \textit{backward propagation} to generate gradient tensors, which can be aggregated synchronously or
asynchronously among GPUs. 
Synchronous data-parallel DDL, where all GPUs communicate the gradient tensors and wait for the aggregated results prior to the next iteration, is the de facto standard used by DDL frameworks~\cite{byteps, pytorch_ddp, horovod, tensorflow}; asynchronous data-parallel DDL, where GPUs do not wait for aggregation to complete, can hurt the model accuracy~\cite{chen2016revisiting}.
We focus on synchronous data-parallel DDL because of its wide adoption.

\begin{figure}[t!]
\begin{center}
\includegraphics[width=0.8\linewidth]{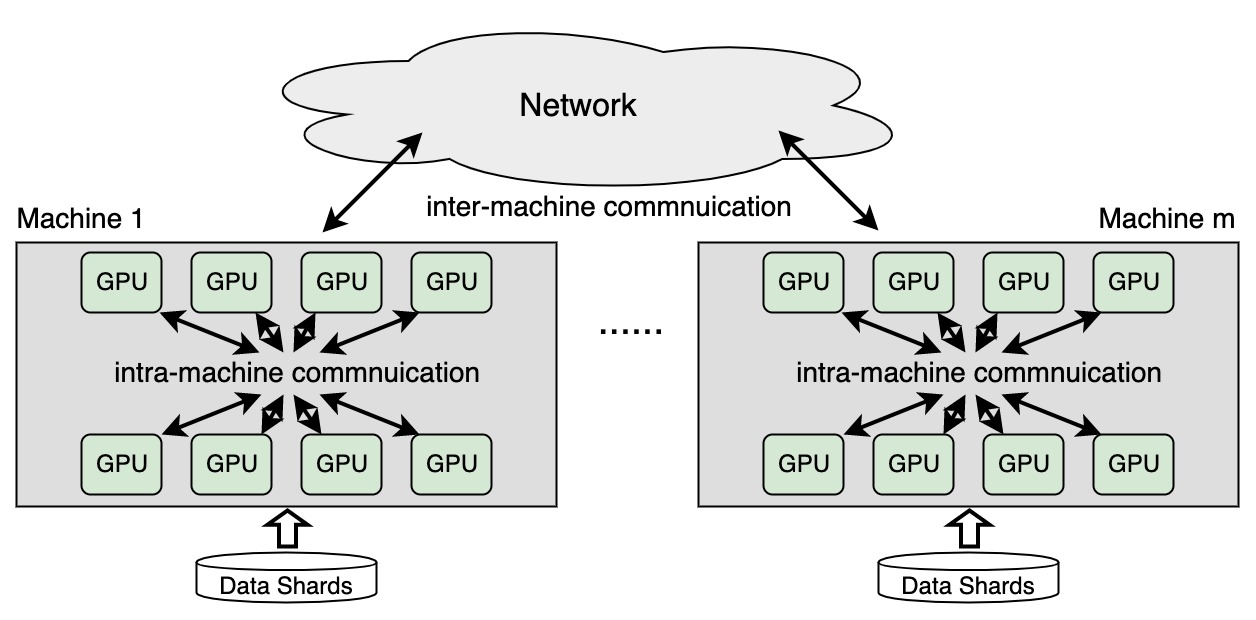}
\caption{Hierarchical communication in DDL.}
\label{fig:comm_ddl}
\end{center}
\end{figure}

Because DDL typically employs multiple machines and each machine has multiple GPUs, it involves both intra-machine and inter-machine communication.
\textit{Hierarchical communication} (as shown in Figure~\ref{fig:comm_ddl}) is widely applied in DDL frameworks~\cite{byteps, horovod, pytorch_ddp, mxnet} because the intra-machine network is usually faster than the inter-machine network.
There are three \textit{phases} for gradient synchronization in hierarchical communication:
1) the gradients are first aggregated among GPUs within one machine;
2) they are then aggregated across machines; 
and 3) the aggregated gradients are communicated within one machine again to ensure that all GPUs have the same synchronized results.
\textit{Flat communication}, i.e., all GPUs join the same collective operation and have only one communication phase, is also supported in some frameworks~\cite{horovod, pytorch_ddp}.

\vspace{-0.1in}

\subsection{Computation and Communication Tension}

Because of the layered structure and a layer-by-layer computation pattern in DNN models~\cite{bai2020pipeswitch}, the wait-free back-propagation mechanism (WFBP)~\cite{poseidon, horovod, byteps, pytorch_ddp, mxnet} is widely adopted to overlap communication with computation in DDL to reduce the iteration time.

However, there still exists an exacerbating tension between computation and communication.
The recent advancements in ML hardware accelerators~\cite{nvidia_performance} and specialized software stacks~\cite{chen2018tvm, zheng2020ansor, rotem2018glow} have significantly improved the single-GPU training speed.
For instance, the single-GPU iteration time of ResNet50 has seen a 22$\times$ decrease in the last seven years~\cite{training_time}.
Faster training speed leads to more frequent gradient synchronization and higher demands on the network.
Unfortunately, network upgrades have not kept up with the pace of computation-related advancements.
The network bandwidth in GPU clouds has only seen a roughly $10\times$ increase in the same period~\cite{Mellanox_update, OpenAI, nvidia_performance}.
This imbalance between the fast-growing computing capability and the slower-growing communication bandwidth reduces the chance to overlap communication with computation, and results in poor scalability of DDL.

To illustrate, we trained real-world DNN models on BytePS-0.2.5~\cite{byteps}, a highly-optimized DDL framework, with 64 NVIDIA V100 GPUs (8 GPUs per machine) and a 100Gbps inter-machine Ethernet network. We measure the scaling factor~\cite{zhang2020network, omnireduce}, which is defined as $\frac{T_n}{nT}$, where $T$ is the training throughput of a single device and $T_n$ is the throughput of DDL with $n$ devices.
BytePS only achieves the scaling factors of 0.58 and 0.51 for the training of two representative and popular DNN models, GPT2 and BERT-base, with NVLink 2.0 for GPU-to-GPU interconnection, as shown in Table~\ref{table:scaling}. 
To put this into context, the training time of BERT-base is about 1200 GPU hours under ideal linear scaling~\cite{bert_training_time}, but in practice, it will take 2350 GPUs hours with 64 GPUs due to the communication time caused by gradient synchronization.
Thus, DNN practitioners have to spend nearly twice the amount of money on training because the cost linearly increases with the required GPU hours~\cite{aws_price}.

\begin{table}[t!]
\scriptsize
\centering
    \begin{tabular}{c|cccc}
    \hline
    \\[-1em]
    \pbox{16cm}{Model} & \pbox{20cm}{Networks} & \pbox{20cm}{FP32} & \pbox{20cm}{GC with GPU} & \pbox{20cm}{GC with CPU} \\ \hline \hline
    GPT2 & NVLink, 100Gbps & 0.58 & 0.67 ($+15\%$) &  0.64 ($+10\%$) \\
    BERT-base &  NVLink, 100Gbps & 0.51 & 0.55 ($+8\%$) & 0.61 ($+20\%$)  \\
    LSTM & PCIe, 25Gbps & 0.46 & 0.43 ($-6\%$) &  0.42 ($-9\%$) \\
    \hline
    \end{tabular}
    \caption {The scaling factors of three popular DNN models with 64 GPUs (8 GPUs per machine) and hierarchical communication. FP32 is the training without GC.}
    \label{table:scaling}
    \vspace{-0.35in}
\end{table}

When network bandwidth in GPU clouds has not kept pace with the improvements in computation, an alternative is to shrink the communicated traffic volume by applying gradient compression.

\vspace{-0.1in}

\subsection{Gradient Compression}
Many gradient compression (GC) algorithms have been proposed in the machine learning community.
\textit{Sparsification} and \textit{Quantization} are the two main types of GC algorithms.
Sparsification selects a subset of the original stochastic gradients for synchronization~\cite{stich2018sparsified, aji2017sparse, dgc} and it can save up to 99.9\% of the gradient exchange while maintaining model accuracy~\cite{dgc}.
Quantization decreases the precision of gradients;
gradients in single-precision floating-point format (FP32) are mapped to fewer bits, such as 8 bits~\cite{8bits}, 2 bits~\cite{wen2017terngrad}, and even 1 bit~\cite{1bit, efsignsgd, signsgd} to reduce the communicated traffic volume by up to 96.9\%.
Such compression algorithms have been theoretically proven and/or empirically validated to preserve the convergence of model training and impose negligible impact on model accuracy when combined with error-feedback mechanisms~\cite{1bit, wu2018error, stich2018sparsified, dgc, jiang2018linear}.
The industry is adopting GC because of its great potential to alleviate the communication bottleneck in DDL. 
The efforts from Meta, AWS, and ByteDance to bring GC to mainstream DNN systems have begun recently~\cite{pytorch_gc, aws_gc, zhong2021compressed}.
However, the scalability improvement of DDL via GC has been still poor.









\vspace{-0.15in}

\section{Challenges of Applying GC to DDL}
\label{sec:challenges}
We first define some key terms.

\noindent
$\bullet$ \textbf{Tensor computation} is the computation of a tensor during backward propagation.

\noindent
$\bullet$ \textbf{Communication time} is the wall-clock time for communication. It is denoted as $\tau_{comm}$.

\noindent
$\bullet$ \textbf{Communication overhead} is the communication time that cannot overlap with tensor computation of any tensors. It is denoted as $o_{comm}$.

\noindent
$\bullet$ \textbf{Compression time} is the wall-clock time to perform compression and decompression operations on devices, e.g., GPUs or CPUs. It is denoted as $\tau_{comp}$

\noindent
$\bullet$ \textbf{Compression overhead} is the compression time that cannot overlap with either tensor computation or communication of any tensors. It is denoted as $o_{comp}$.

Although GC can reduce $\tau_{comm}$, its compression overheads can dramatically dilute the benefits gained from the reduced communication time.
To demonstrate this, we apply a popular sparsification algorithm, DGC~\cite{dgc}, to the aforementioned GPT2 training and a representative 1-bit quantization algorithm, EFSignSGD~\cite{efsignsgd}, to BERT-base training.
The compression rate of DGC is 1\%, i.e., only 1\% of gradients are exchanged during synchronization.
Tensors are compressed with GPUs~\cite{HiPress} or CPUs~\cite{zhong2021compressed} in separate experiments.
As shown in Table~\ref{table:scaling}, GC only achieves up to 20\% training speedup, which is on par with the findings in prior works~\cite{xu2021grace, HiPress, agarwal2021utility}.
In fact, GC can harm performance in some situations. To illustrate, we apply DGC with 1\% compression rate to the training of LSTM~\cite{LSTM} on 64 V100 GPUs with PCIe 3.0 x16 as the intra-machine network and 25Gbps inter-machine Ethernet.\footnote{NVLink 2.0 gives every GPU in total 1.2Tbps GPU-GPU bandwidth, but PCIe 3.0 x16 only provides $\sim$100Gbps bandwidth~\cite{byteps}. PCIe-only GPU machines are common in GPU clusters that have 25Gbps Ethernet~\cite{byteps, alibaba_gpu_cloud, google_cloud, ranganath2021mapa}.}
As listed in Table~\ref{table:scaling}, GC slows down training by up to 9\%.

\begin{figure}[t!]
\begin{center}
\includegraphics[width=1\linewidth]{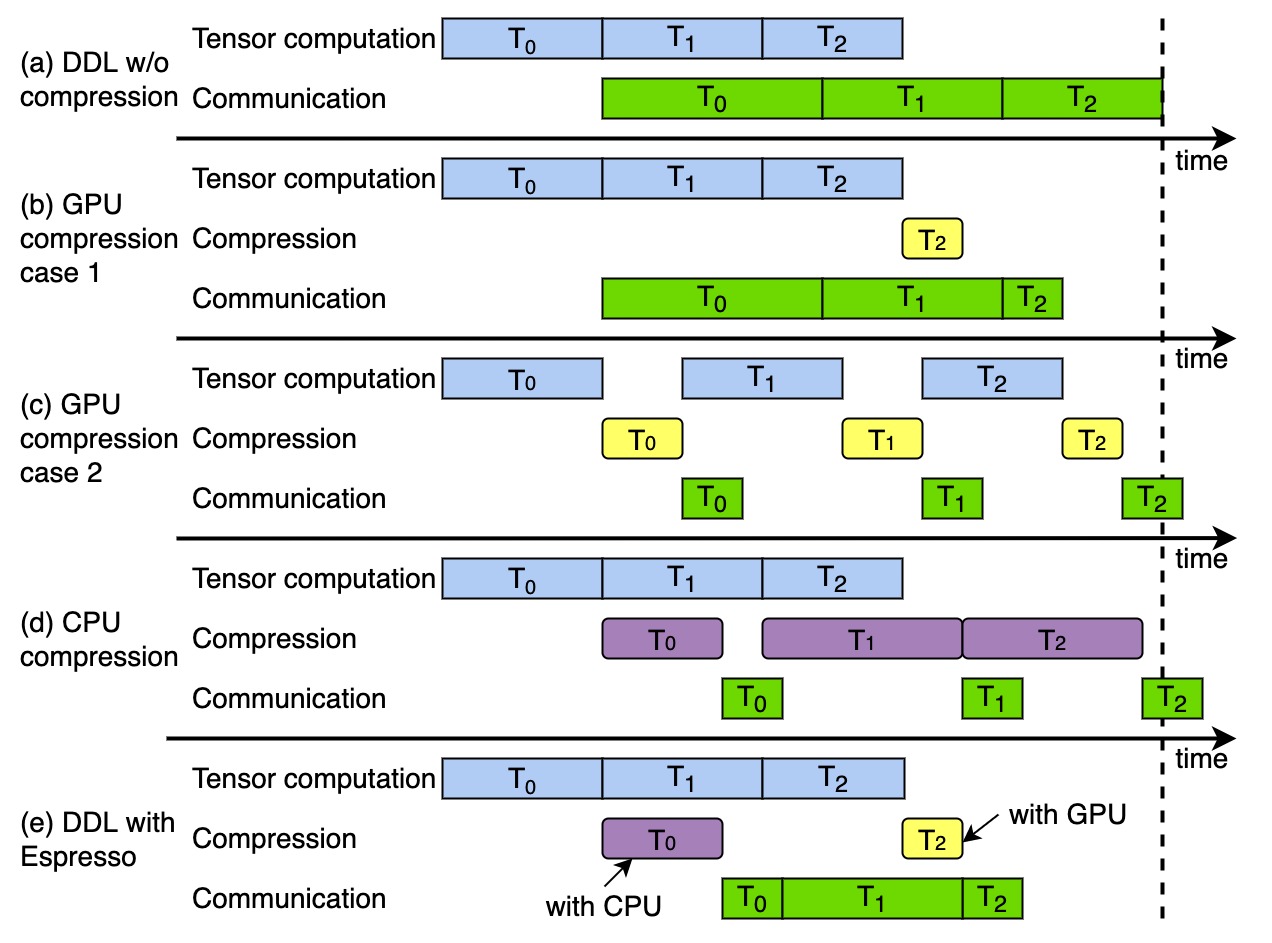}
\vskip -0.2in
\caption{A DDL example with different compression strategies. 
(a) is the baseline; (b) reduces the iteration time, but it is not optimal; (c) and (d) harm the performance; (e) is our solution and achieves optimal performance.
The communication and compression overheads depend on the interactions among tensors. The decompression operations are omitted.}
\vspace{-0.2in}
\label{fig:overhead_time}
\end{center}
\end{figure}

In the following, we will explain the root reasons why it is challenging to obtain large benefits from GC for DDL.

\vspace{-0.1
in}

\subsection{Root Reasons of the Challenges}
\label{sec:reasons}

The choice of compression strategies determines the iteration time of compression-enabled DDL.
Figure~\ref{fig:overhead_time} is an example that shows the timelines of tensor computation, communication, and compression of DDL with different compression strategies.
Figures~\ref{fig:overhead_time}(a) is the baseline without GC and it illustrates the tensor computation time (blue boxes) and communication time (green boxes) of all tensors, i.e., T$_0$, T$_1$, and T$_2$.
Figure~\ref{fig:overhead_time}(b) compresses T$_2$ with GPUs and it reduces the iteration time.
Figures~\ref{fig:overhead_time}(c) and (d) compress the three tensors with GPUs and CPUs, respectively, but unfortunately, they both harm the performance of DDL.
Figures~\ref{fig:overhead_time}(e) shows the optimal compression strategy with {\name}.

It is challenging to find the optimal compression strategy.
Applying GC to DDL is essentially to reduce the communication overheads at the cost of the compression overheads.
The optimal compression strategy maximizes the difference between the reduced communication overheads and the incurred compression overheads.
There are three root reasons for the challenges.

\textbf{Reason \#1.}
\textit{It is hard to quantify the communication and compression overheads because of the intricate interactions among tensors.}

\textbf{Communication may or may not overlap with tensor computation.}
The overlapping time of different tensors can vary.
For example, in Figure~\ref{fig:overhead_time}(a), T$_0$'s $o_{comm}$ is zero because its communication is fully overlapped with tensor computation, but T$_2$'s $o_{comm}$ is its communication time because it has no overlap with tensor computation.
Moreover, the overlapping time of one tensor can vary under different compression strategies.
For example, in Figure~\ref{fig:overhead_time}(a), T$_1$'s communication partially overlaps with T$_2$'s tensor computation. 
However, in Figure~\ref{fig:overhead_time}(c), after compression, T$_1$'s communication can completely overlap with T$_2$'s tensor computation. Furthermore, in Figure~\ref{fig:overhead_time}(d), T$_1$'s communication has no overlap with the computation of other tensors. 
Hence, it is difficult to quantify the communication overhead of each tensor.

\textbf{Compression may or may not overlap with tensor computation and communication.}
How much $\tau_{comp}$ can be overlapped highly depends on the strategy.
For instance, in Figure~\ref{fig:overhead_time}(b), T$_2$'s GPU compression fully overlaps with T$_1$'s communication.
In Figure~\ref{fig:overhead_time}(d), T$_1$'s CPU compression partially overlaps with T$_2$'s tensor computation.
In Figure~\ref{fig:overhead_time}(c), the three GPU compressions are fully exposed.
Hence, it is difficult to quantify the compression overhead.

\textbf{Only considering $\tau_{comm}$ and $\tau_{comp}$ for the decision of compression strategies can harm the performance.}
Figure~\ref{fig:overhead_time}(c) maximizes the difference between the reduced communication time and the compression time by compressing the three tensors.
However, because GPU compression competes for compute resources with tensor computation, it delays training and prolongs the iteration time instead.
Hence, we must consider $o_{comm}$ and $o_{comp}$ to determine compression strategies for compression-enabled DDL.

\begin{figure}[t!]
\begin{center}
\includegraphics[width=0.85\linewidth]{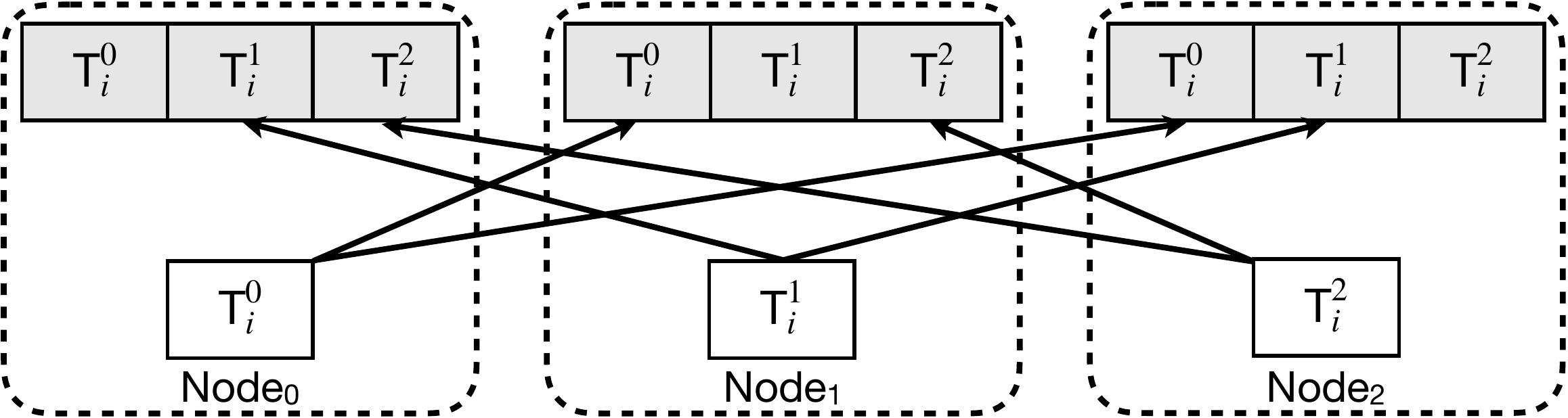}
\vskip -0.15in
\caption{An indivisible scheme. T$_i^j$ is the tensor T$_i$ on Node $j$. Each node retrieves tensors from other nodes.}
\label{fig:allgather}
\end{center}
\vskip -0.3in
\end{figure}

\begin{figure}[t!]
\begin{center}
\includegraphics[width=1.0\linewidth]{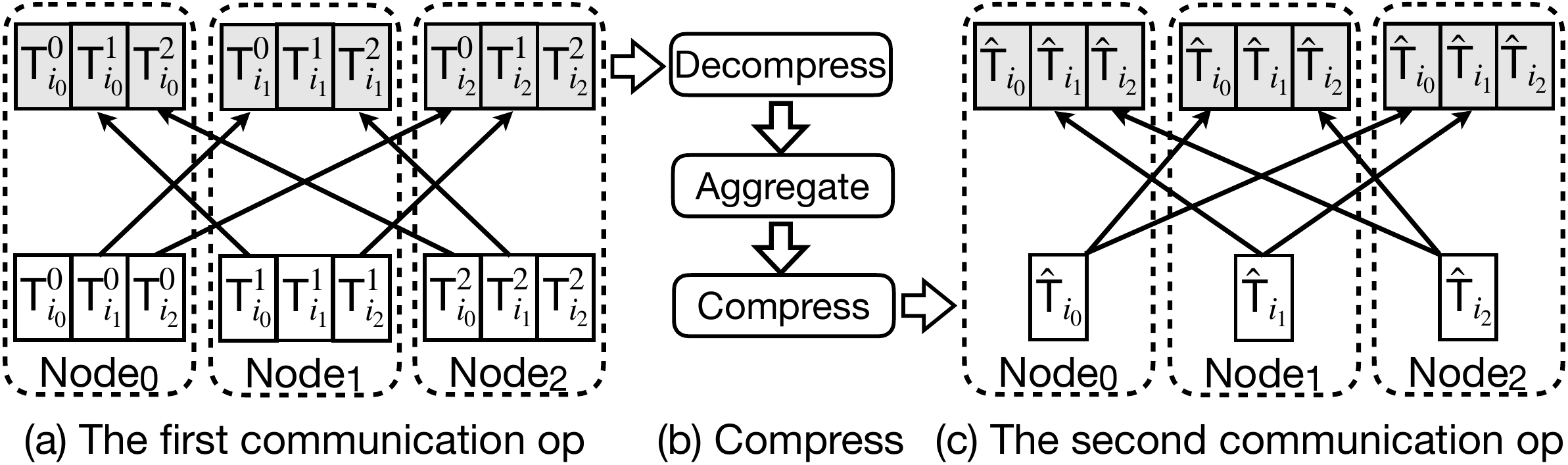}
\vskip -0.15in
\caption{A divisible scheme. In (a), T$_i^j$ is partitioned into $3$ parts, i.e., T$_{i_0}^j$, T$_{i_1}^j$, and T$_{i_2}^j$. The first communication operation (op) is a shuffle. After Node $j$ receives the $j_{th}$ part from other nodes, it decompresses and aggregates them.
It then compresses the aggregated tensor and obtains $\hat T_{i_j}$ for the second communication op.}
\label{fig:alltoall}
\end{center}
\vskip -0.2in
\end{figure}



\begin{figure*}[t!]
\centering
  \includegraphics[width=1.03\linewidth]{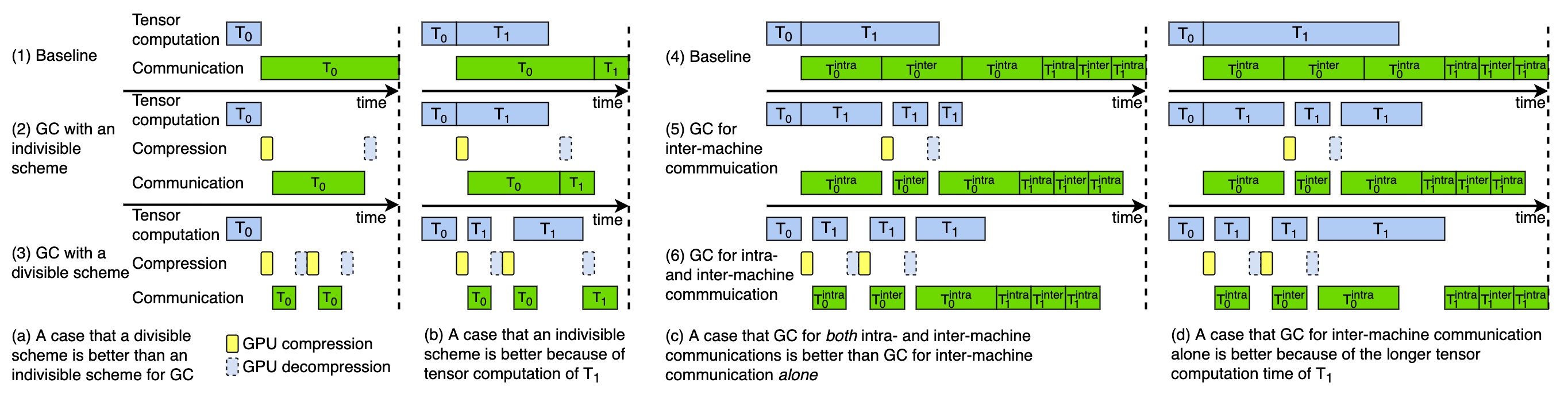}
  \vskip -0.15in
  \captionof{figure}{(a) and (b) show that the choice of communication schemes depends on the interactions among tensors. Only T$_0$ is compressed. (c) and (d) show that the decision to apply GC to inter-machine communication alone or to both intra- and inter-machine communications also depends on the interactions among tensors. 
  T$_i^\text{intra}$ and T$_i^\text{inter}$ are T$_i$'s intra- and inter-machine communications.}
  \label{fig:dimension23}
  \vskip -0.15in
\end{figure*}

\textbf{Reason \#2.}
\textit{It is hard to choose the right communication schemes for compressed tensors because of Reason \#1.}

\textbf{There are two types of communication schemes for compressed tensors}:  \textit{indivisible schemes} and \textit{divisible schemes}.
We first consider the case that there are $N$ machines in DDL and each machine has a single GPU.
An indivisible scheme has only one communication operation, as shown in Figure~\ref{fig:allgather}.
Once a tensor is compressed, each node (e.g., GPU or CPU) broadcasts its compressed tensor to other nodes.
After communication, each node decompresses these compressed tensors and aggregates them.
In contrast, a divisible scheme has two communication operations, as shown in Figure~\ref{fig:alltoall}.
Tensors are first compressed and partitioned into $n$ parts, where $1 \le n \le N$. 
The $j_{th}$ node receives the $j_{th}$ part from other nodes.
It then performs decompression, aggregation, and the second compression operation.\footnote{In some cases it can skip these three operations to begin the next communication directly.}
After that, it broadcasts the compressed tensor to other nodes.
After communication, each node decompresses these compressed tensors and aggregates them.

\textbf{It is hard to decide between indivisible and divisible schemes for GC.}
Compared to indivisible schemes, divisible schemes have lower communication time and higher compression time due to the two compression and decompression operations.
As shown in Figures~\ref{fig:dimension23}(a), GC with a divisible scheme outperforms GC with an indivisible scheme.
However, in Figure~\ref{fig:dimension23}(b), T$_0$'s communication overlaps with T$_1$'s tensor computation and an indivisible scheme outperforms a divisible scheme for GC.
Thus, the decision of communication schemes depends on the interactions among tensors.

\textbf{Reason \#3.}
\textit{It is hard to determine whether to apply GC to intra- or inter-machine communication or both to alleviate communication bottleneck because of Reasons \#1 and \#2.}

\textbf{DDL can involve both intra- and inter-machine communications.}
We now consider the case that there are $N$ machines and each machine has $k$ GPU, where $k>1$, as shown in Figure~\ref{fig:comm_ddl}.
It has intra- and inter-machine communications, and both can become the performance bottleneck.

\textbf{Whether to apply GC to intra- or inter-machine communication or both depends on the interactions among tensors.}
If a tensor is only compressed for inter-machine communication, intra-machine communication can still be a performance issue.
Figure~\ref{fig:dimension23}(c) shows that applying GC to intra-machine communication can further reduce the iteration time.
However, if T$_1$ has a longer computation time, it can overlap more time with T$_0$'s communication, as shown in Figure~\ref{fig:dimension23}(d). 
In this case, applying GC to both intra- and inter-machine communications leads to worse performance than applying it to inter-machine communication alone.

\textbf{This decision also depends on the chosen communication schemes.}
Because both intra- and inter-machine communications need to choose from indivisible or divisible schemes, the difficulties to determine the right schemes make the decision of the compression choices even harder.

\vspace{-0.16in}

\subsection{Research Questions}
In light of the three root reasons, there are two research questions to answer for applying GC to DDL.


\textbf{Question \#1: how to express all possible compression strategies and interactions among tensors for DDL regardless of different distributions of computation and communication time of tensors in different DNN models, different intra- and inter-machine bandwidth, and different GC algorithms?}

Applying GC to a tensor must answer the following fundamental questions: 
Does it need compression? 
If so, what type of compute resources to use for its compression?
After compression, what communication schemes should the compressed tensor use?
If it has multiple communication phases, where to compress and decompress this tensor?
The search space is huge when holistically considering these decisions.
Moreover, there are typically a large number of tensors in a DNN model and the compression option of one tensor can impact the choices of other tensors because of their intricate interactions.
The compression strategy determines the interactions among tensors, which determine the training throughput of compression-enabled DDL.
Therefore, it is crucial to express all the strategies and the corresponding interactions among tensors to avoid missing the opportunity to maximize the training throughput.

\textbf{Question \#2: how to analyze the interactions among tensor computation, communication, and compression, as well as the different performance trade-offs of GPUs and CPUs for GC, to determine a near-optimal compression strategy for DDL and to do so quickly?}

Even if all compression strategies are at hand, the time complexity to find the optimal strategy is exponential (\S\ref{sec:problem}).
The searching time can be much longer than the training time, which is unacceptable.
Moreover, the optimal strategy is specific to each situation depending on the DNN model, intra- and inter-machine bandwidth, GC algorithm, etc., and thus cannot be reused across situations. 
A successful solution to this question must develop new insights on the interactions among tensors and the different performance trade-offs with different types of compute resources for GC that can eliminate suboptimal strategies from consideration.



\vspace{-0.1in}
\section{The Design of \name}


\subsection{Overview}
To maximize the training throughput of compression-enabled DDL, the core idea of {\name} is to select a near-optimal compression strategy from an extremely large search space with the following two techniques.




\noindent
\textbf{A decision tree abstraction to describe all compression options of any tensors as well as empirical models for the time of tensor computation, communication, and compression to express all compression strategies and interactions among tensors.}
The abstraction can express all types of compute resources for compression, communication schemes, and different choices to apply GC to intra- and inter-machine communications. 
It can also serve as the building block to describe all the compression strategies of any compression-enabled DDL.
The empirical models enable {\name} to derive the timeline of tensor computation, communication, and compression of all tensors, and thus their intricate interactions with any compression strategy.


\noindent
\textbf{An algorithm for selecting a near-optimal compression strategy with four properties.}
The algorithm
1) rules out tensors that certainly bring no benefits to DDL with GC based on the analysis of interactions among tensors;
2) uses a prioritization method for applying GC to tensors to maximize the benefits with the minimum number of tensors for compression;
3) determines the compression options with the compression and communication overheads based on the analysis of interactions, rather than with the wall-clock time; 
and 4) finds a provably optimal solution to offload compression from GPUs to CPUs.

\vspace{-0.1in}
\subsection{The Decision Tree Abstraction}

\subsubsection{The dimensions of the search space}
\label{sec:dimension}
\begin{figure}[t!]
\begin{center}
\includegraphics[width=\linewidth]{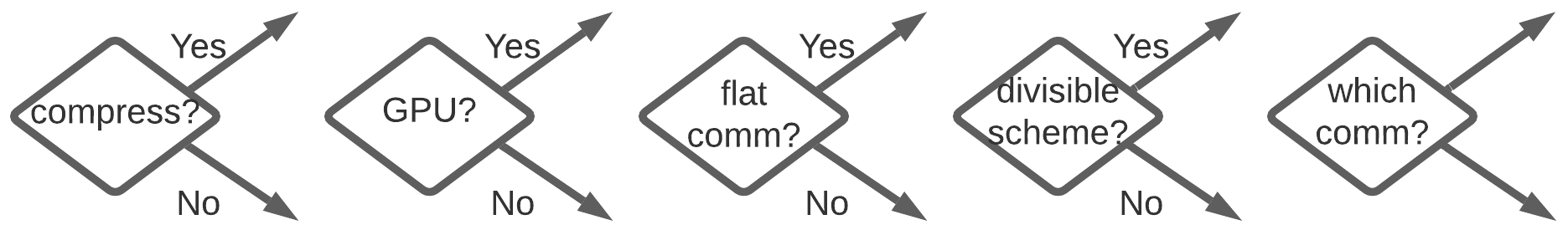}
\vskip -0.1in
\caption{The decision tasks. \textit{compress?} is for Dimension 1, \textit{GPU?} is for Dimension 2, and the other three decision tasks are for Dimension 3. The options of Dimension 4 is illustrated in Figure~\ref{fig:tree}.}
\label{fig:decision_task}
\end{center}
\vspace{-0.2in}
\end{figure}

There are four dimensions that {\name} must consider to describe the search space of compression options for each tensor.
The decision tasks for these dimensions are shown in Figure~\ref{fig:decision_task}.

\noindent \textbf{Dimension 1: compression or no compression.}
Because GC can incur non-negligible compression time and even harm performance, there is no need to compress all tensors.
{\name} must determine the set of tensors that should be compressed to maximize the benefits of GC.

\noindent \textbf{Dimension 2: GPU or CPU for compression.}
Both GPUs and CPUs can be used for GC to minimize the compression overhead.
{\name} must determine the set of tensors in a DNN model for GPU and CPU compression, respectively.
Task Comp and Task Decomp, as listed in Table~\ref{table:action_task}, are the action tasks to decide between GPUs and CPUs for compression and decompression operations, respectively.

\begin{table}[t!]
    \scriptsize
    \centering
    \begin{tabular}{l|ll}
    \hline
    Routines  & Uncompressed tensors   & Compressed tensors    \\ \hline \hline
    Indivisible schemes & Allreduce      & Allgather                              \\[0.2cm]
    Divisible schemes   & \pbox{20cm}{Reduce-scatter/Allgather \\Reduce/Broadcast}  & \pbox{20cm}{Alltoall/Allgather \\Gather/Broadcast} \\[0.2cm] \hline
    \end{tabular}
    \caption {The collective routines for synchronization.}
    \vspace{-0.25in}
    \label{table:routines}
\end{table}

\noindent \textbf{Dimension 3: the communication schemes.}
Compressed tensors cannot use Allreduce for synchronization because their aggregation operations are not associative~\cite{agarwal2021utility, HiPress, xu2021grace}.
Both indivisible and divisible communication schemes can be used, while each can have more than one choice of collective routines, i.e., one collective communication operation or an operation pair.
Table~\ref{table:routines} lists the common collective routines used in DDL for GC~\cite{MPIoptimization, NCCL, pytorch_ddp}.
Because tensors can be communicated without GC, Table~\ref{table:routines} lists the collective routines for uncompressed tensors as well.
We distinguish the two communication operations in a divisible scheme as its first and second steps.
In addition, flat and hierarchical communications lead to a different number of communication phases for gradient synchronization.
Therefore, this dimension requires {\name} to consider three sub-dimensions: flat or hierarchical communication, indivisible or divisible schemes, and specific collective routines for each communication phase.
The decision tasks of the three sub-dimensions are shown in Figure~\ref{fig:decision_task} as \textit{flat comm?}, \textit{divisible scheme?}, and \textit{which comm?}.
Because both uncompressed and compressed tensors have indivisible and divisible schemes, and division schemes have two collective operations,
\textit{which comm?} has six action tasks, as listed in Table~\ref{table:action_task}.

\begin{figure*}[t!]
\begin{center}
\includegraphics[trim=5 5 2 5, clip, width=0.9\linewidth, height=2.8in]{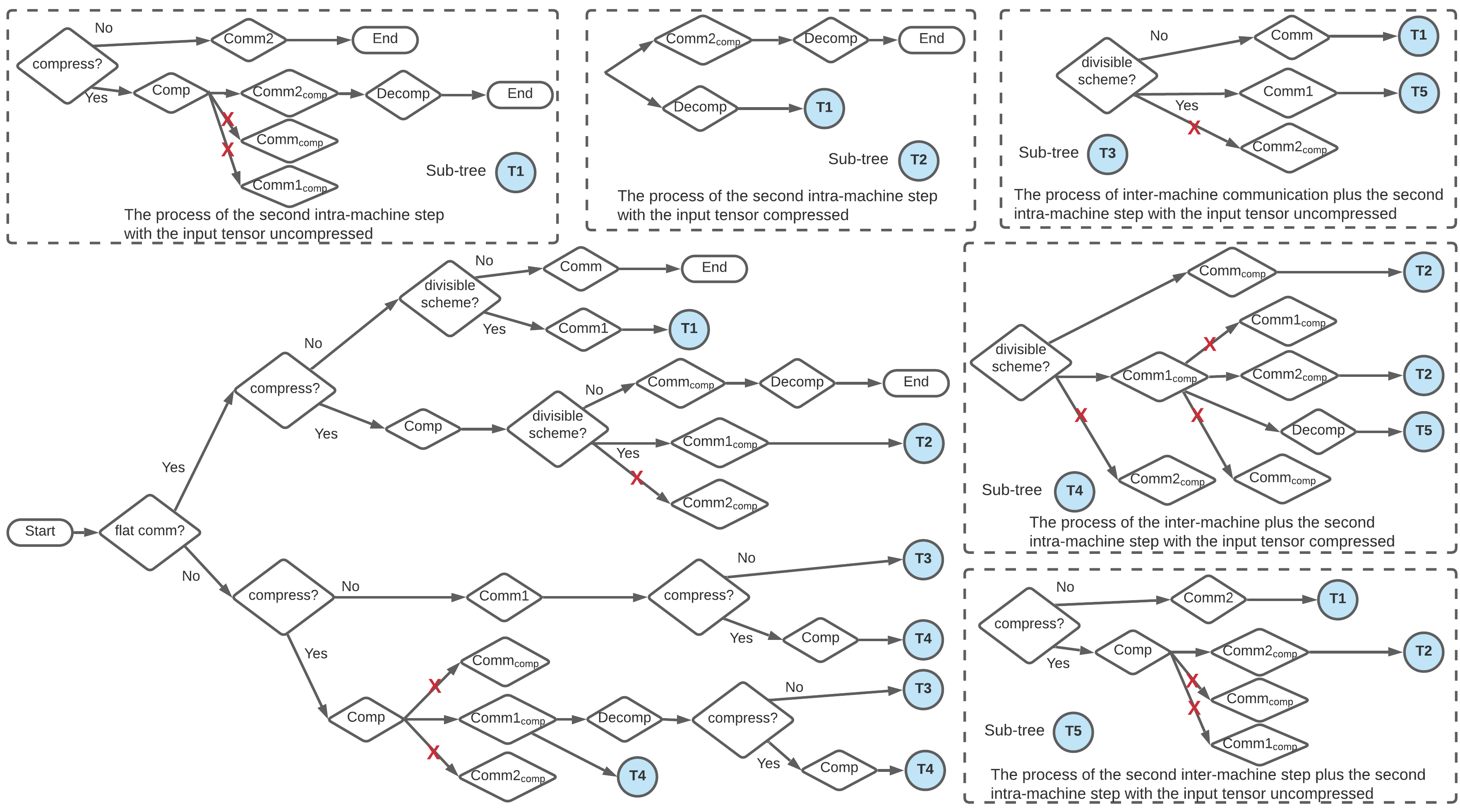}
\vskip -0.15in
\caption{The decision tree abstraction for the compression options. Each diamond in the tree is a decision task.} 
\vskip -0.2in
\label{fig:tree}
\end{center}
\end{figure*}

\noindent
\textbf{Dimension 4: the compression choice.}
It determines where to perform compression and decompression operations.
For flat communication, it has two \textit{communication patterns} because it can choose from an indivisible or a divisible scheme.
For hierarchical communication, it can choose from a divisible or an indivisible scheme for its inter-machine communication.
Although it can also choose from a division scheme or two indivisible schemes for its two intra-machine communications, the former is better than the latter due to the less amount of traffic volume.
Therefore, {\name} only considers division schemes for intra-machine communications in hierarchical communication.
Tensors can be compressed as long as they need communication and compressed tensors can be decompressed after any communication operation. 
All the options for this dimension, i.e., the possible positions of Task Comp and Task Decomp in each communication pattern, are illustrated in Figure~\ref{fig:tree}.

\begin{table}[t!]
\centering
\scriptsize
    \begin{tabular}{l|ll}
    \hline
    Action Tasks    & Description       & Search space                                               \\ \hline \hline
    Comp            & Compression operation                                 & \{CPU, GPU\}                                              \\ \hline
    Decomp          & Decompression operation                               & \{CPU, GPU\}                                              \\ \hline
    Comm            & Indivisible scheme for UT                            & \{Allreduce\}                                             \\ \hline
    Comm1           & The first step of a DS for UT   & \{Reduce-scatter, Reduce\}                                \\ \hline
    Comm2           & The second step of a DS for UT  & \{Allgather, Broadcast\}                                  \\ \hline
    Comm$_{comp}$  & Indivisible scheme for CT                        & \{Allgather\}                     \\ \hline
    Comm1$_{comp}$ & The first step of a DS for CT     &  \{Alltoall, Gather\} \\ \hline
    Comm2$_{comp}$ & The second step of a DS for CT    & \{Allgather, Broadcast\} \\ \hline
    \end{tabular}
    \caption {The eight action tasks. UT denotes uncompressed tensors, CT denotes compressed tensors, and DS denotes divisible schemes.}
    \vskip -0.3in
    \label{table:action_task}
\end{table}

\vspace{-0.1in}

\subsubsection{Constructing the tree}
\label{sec:construct}


A \textit{compression option} is a series of decision tasks that determine all the communication and compression operations of a tensor for its synchronization.
These operations have orders and dependencies.
There are eight action tasks (as listed in Table~\ref{table:action_task}), but not all of them can have direct connections, i.e., a task is performed right after another.
The valid connections of action tasks are omitted due to space limitations.

\noindent\textbf{Tree construction.}
Based on the four dimensions and the valid connections of the eight action tasks, {\name} can express all possible compression options of any tensor with a decision tree, as shown in Figure~\ref{fig:tree}.
Because the choices of GPUs or CPUs for Task Comp and Task Decomp do not impact communication tasks, we use one arrow to represent their two choices for simplicity.


There are three \textit{pruning rules} to construct the tree.
The first rule is that the following action tasks of an action task must be its valid connections.
The second rule is that the communication tasks must match the correct steps.
For example, Comm1 and Comm1$_{comp}$ are only valid as the first steps of divisible schemes.
The third rule is that the choices of communication tasks in the first and second steps must pair.
For example, if Comm1 is Alltoall, then Comm2 in this divisible scheme must be Allgather.
Each path from Start to End is a valid compression option.
The red crosses in Figure~\ref{fig:tree} are the invalid paths ruled out by these pruning rules.

There are five sub-trees illustrated in Figure~\ref{fig:tree} to abstract parts of the tree.
Sub-tree $T_1$ and $T_2$ describe the process of the second intra-machine step with the input tensor uncompressed and compressed, respectively.
Sub-tree $T_3$ and $T_4$ describe the process of inter-machine communication plus the second intra-machine step with the input tensor uncompressed and compressed, respectively.
Sub-tree $T_5$ describes the process of the second inter-machine step plus the second intra-machine step with the input tensor uncompressed.


\noindent
\textbf{Expressiveness and extensibility.}
Because all the valid connections between decision tasks have been considered, this decision tree abstraction can cover all the possible compression options. 
It is easy for {\name} to extend the search space of communication tasks to consider new communication schemes for GC~\cite{sparcml, omnireduce} and other types of compute resources~\cite{tpu, fpga}.
In addition, it allows users to manually add constraints to prune the decision tree to rule out undesirable compression options for their applications. 
For example, users can limit the number of compression operations for each tensor to avoid the accuracy loss of training models.

\noindent
\textbf{Compression strategies.}
Let $\mathcal{T} = \{T_i\}$ denote the set of tensors in a DNN model and the number of tensors in $\mathcal{T}$ is $|\mathcal{T}| = N$.
$\mathcal{C}$ is the set of all possible compression options.
$S = \{c_j\}$ is a compression strategy for the DNN model, where $c_j \in \mathcal{C}$ is the compression option for tensor $T_j$.

\vspace{-0.2in}
\subsection{Empirical interactions among tensors}
\label{sec:interaction}
The decision tree abstraction can express all compression strategies, but it is incapable of describing the intricate interactions among tensors, which determine the choice of compression strategies for different DDL training jobs.
To describe the interactions, {\name} proposes different methods to empirically model the time of tensor computation, communication, and compression, respectively.


\noindent
\textbf{Tensor computation.}
{\name} needs the computation time of each tensor.
It collects execution traces of DNN training jobs without GC for 100 iterations to capture the starting and ending time of the computation of each tensor during backward propagation. 
{\name} then averages the computation time.
It also collects the information of tensor sizes.

\noindent
\textbf{Communication time.} 
{\name} needs the communication time of tensors with and without GC.
Given a tensor, {\name} predicts its communication time with different communication schemes and network bandwidth.
The cost models follow the model analysis in the literature~\cite{MPIoptimization, broadcast}.
These communication models account for different tensor sizes, communication schemes, the number of machines and GPUs, and network bandwidth.

\noindent
\textbf{Compression time.} 
{\name} also predicts the compression time of tensors with different sizes and different types of compute resources.
Based on the information collected from execution traces, it can have all the possible tensor sizes as the input of compression and decompression operations.
For any GC algorithm, {\name} profiles its computational time of these operations on GPUs and CPUs, respectively. 
It runs compression and decompression operations with different tensor sizes 100 times and then averages the results.
We observe that both the tensor computation time and the compression time keep almost constant across runs~\cite{zhang2020network, training_time}.

\noindent
\textbf{Expressing interactions.} 
Given these empirical models and a compression strategy, {\name} can derive the timeline of tensor computation, communication, and compression of all tensors in a DNN model.
Several timeline examples are shown in Figure~\ref{fig:overhead_time}.
It can obtain the overlapping time of tensors and thus their interactions based on the timeline.

In the next section, {\name} will exploit the timeline and analyze the interactions among tensors to obtain a near-optimal compression strategy for compression-enabled DDL.

\vspace{-0.1in}
\subsection{{\name}'s Decision Algorithm}
\label{sec:algo}

\subsubsection{The optimization problem}
\label{sec:problem}

We define the optimization problem as follows to search for the optimal compression strategy for a DDL training job.

\vskip -0.2in
\begin{problem}
    Given a DDL training job and a compression algorithm, how to maximize its training throughput with an optimal compression strategy?
\end{problem}
\vskip -0.1in

Let $F(S)$ be the iteration time with compression strategy $S$.
The objective is to minimize $F(S)$ with the optimal compression strategy.
The difficulty of the problem results from the overlapping time among tensor computation, communication, and compression.
Given a compression strategy, {\name} can obtain the overlapping time of each tensor with other tensors.
However, both CPU and GPU compression delay communications and change the overlapping time accordingly. 
Naively, we can enumerate all possible combinations to find the optimal solution. 
This is not acceptable because the time complexity is $O(|\mathcal{C}|^N)$, where $N$ could be a few hundred and $|\mathcal{C}|$ is 4341 based on the decision tree abstraction in Figure~\ref{fig:tree}.

\begin{figure}[t!]
\begin{center}
\includegraphics[width=0.9\linewidth]{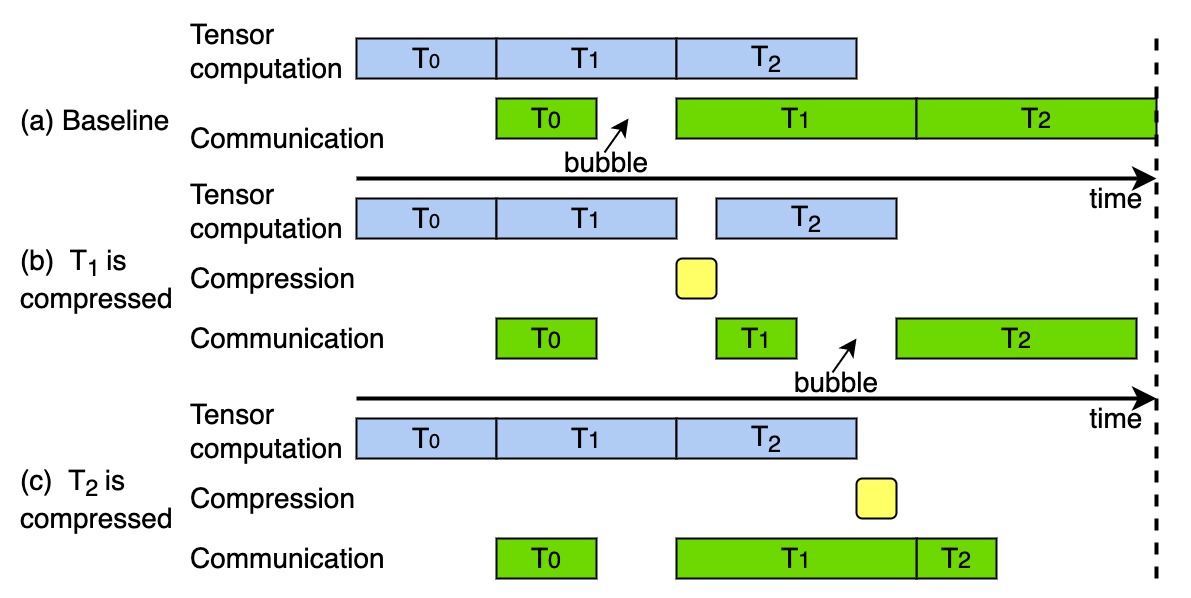}
\vspace{-0.1in}
\caption{(a) shows that tensors communicated before bubbles need no compression. T$_1$ and T$_2$ have the same size. In (b), T$_1$ is compressed and a new bubble is formed. In (c), T$_2$ is compressed and it reduces more iteration time than compressing T$_1$.}
\vspace{-0.2in}
\label{fig:bubble}
\end{center}
\end{figure}

\subsubsection{{\name}'s GPU compression}

To quickly determine a near-optimal compression strategy for DDL, {\name} first considers GPU resources for GC and then offloads compression to CPUs to minimize the contention with tensor computation.
There are three properties for the design of {\name}'s GPU compression decision algorithm.

\textbf{Property \#1.} 
The communication timeline of a DNN model can have \textit{bubbles}, i.e., the gaps between communications of adjacent tensors.
In Figure~\ref{fig:bubble}(a), there is a bubble between the communications of T$_0$ and T$_1$  because T$_1$ is not ready for communication when T$_0$'s communication completes.
There is no benefit to compressing tensors communicated before bubbles because reducing their communication time only widens the gaps, rather than shifts communications of tensors after bubbles to an earlier time.
Compressing these tensors even harms the performance of DDL because of the resource contentions with tensor computation.
We observe that half of the tensors are communicated before bubbles in the training of LSTM with 8 NVLink-based GPU machines in a 100Gbps network.
Moreover, compressing particular tensors can also lead to new bubbles being formed due to the reduced communication time. 
For example, Figure~\ref{fig:bubble}(b) shows that a new bubble appears when T$_2$ is compressed.
Therefore, {\name} rules out uncompressed tensors communicated before bubbles for GC whenever the bubbles appear.


\begin{figure}[t!]
\centering
\begin{minipage}{.50\linewidth}
  \flushleft
  \includegraphics[width=\linewidth]{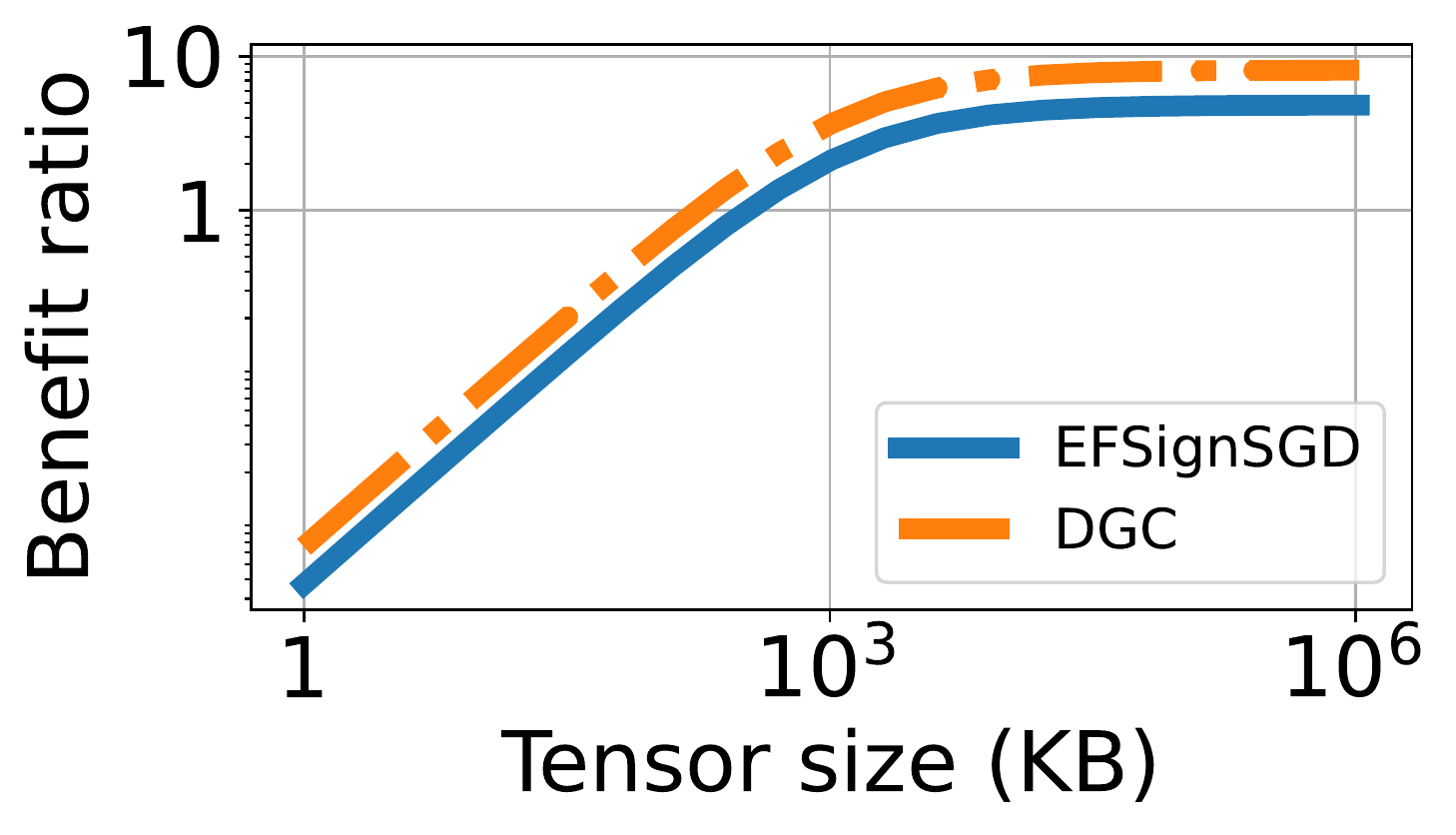}
  \vskip -0.15in
  \captionof{figure}{The benefit ratio of GPU compression.}
  \label{fig:speedup}
\end{minipage}\hspace{\fill}%
\begin{minipage}{.47\linewidth}
  \flushright
  \includegraphics[width=\linewidth]{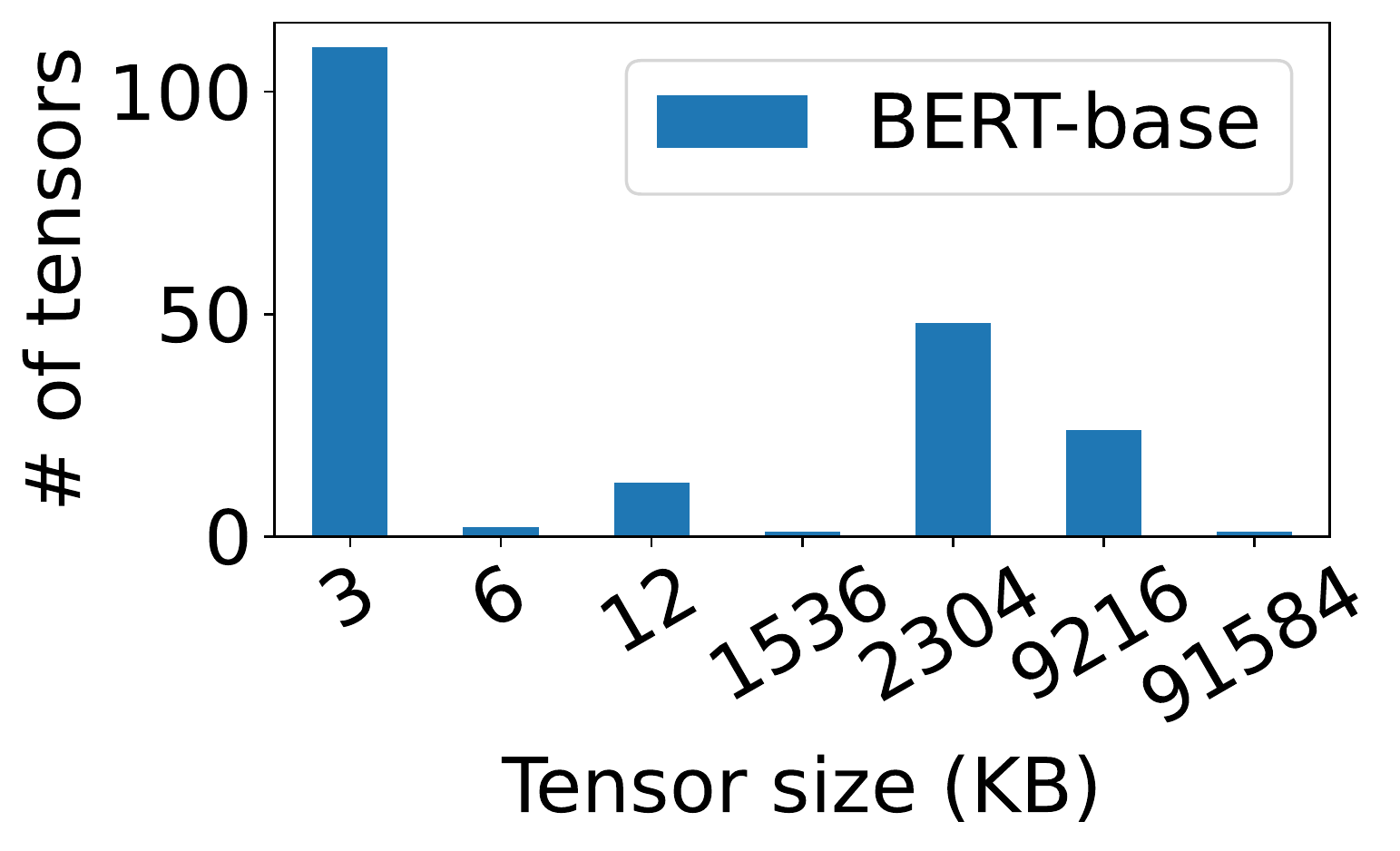}
  \vskip -0.15in
  \captionof{figure}{Number of tensors with the same sizes.}
  \label{fig:tensor_size}
\end{minipage}
\end{figure}

\textbf{Property \#2.} 
There are two insights for the compression order of tensors.
The first one is that compressing larger tensors can bring more benefits to DDL because GC incurs a constant overhead to launch GPU kernels for compression~\cite{wang2021mergecomp, horovod}.
Figure~\ref{fig:speedup} shows the ratio of the reduced communication time to the incurred compression time with 64 GPUs and NVLink.
The ratio increases with tensor sizes and it indicates that GPU compression is more efficient for larger tensors.
The second one is that compressing tensors closer to the output layer, i.e., the last layer during backward propagation, can bring more benefits.
For example, in Figure~\ref{fig:bubble}(c), T$_1$ and T$_2$ have the same size. 
Compressing T$_2$ can reduce more iteration time than compression T$_1$ for two reasons:
1) T$_2$'s compression overlaps more with communication and has no contention with tensor computation,
and 2) compressing T$_2$ can reduce more communication overhead because its communication overlaps less with tensor computation.
Based on these two insights, {\name} applies GC to tensors in the descending order of their sizes, and prioritizes tensors closer to the output layer when they have the same size.


\textbf{Property \#3.}
{\name} considers the communication and compression overheads to determine the compression options.
As discussed in Section~\ref{sec:reasons}, only considering the communication and compression time for the decisions can harm the performance because they can be overlapped with other operations.
Given a tensor, {\name} enumerates the possible compression options and expresses the corresponding interactions among tensors.
It then chooses the one which minimizes the iteration time as the compression option.

\begin{algorithm}[t!]
\footnotesize
\caption{{\name} with GPU compression}
\label{algo:gpu}
\SetKwInput{KwInput}{Input}                
\SetKwInput{KwOutput}{Output}              
\DontPrintSemicolon 
  
    \KwInput{$S$ is a compression strategy and $S[i]$ is the compression option for $T_{i}$. It is initialized with no compression for all tensors.
    $\mathcal{C}_{gpu}$ is the set of all compression options with GPUs only. $G_{m_i}$ is a group of tensors with size ${m_i}$.
    }
    \KwOutput{S}
    
    \SetKwFunction{FMain}{Main}
    \SetKwFunction{FRemove}{Remove}
    \SetKwFunction{FSum}{GetBestOption}

    \SetKwProg{Fn}{Function}{:}{}
    \Fn{\FMain{}}{
        sort all tensors in descending order of their sizes and group them based on their sizes to have $\mathcal{G} = \{G_{m_1}, G_{m_2}, \cdots, G_{m_n}\}$, where $m_1 > \cdots > m_n$ \;
        sort tensors in each group of $\mathcal{G}$ in ascending order of their distances to the output layer of the DNN model \;
        $\text{Remove}(S, \mathcal{G})$ \;
        \For{$i\gets1$ \KwTo $n$}{
            \ForEach{$T_{j} \in G_{m_i}$}{
                \tcp{$S$ is updated after GetBestOption()}
                $S = \text{GetBestOption}(S, j)$ \;
                $\text{Remove}(S, \mathcal{G})$ \;
            }
        }
        \KwRet $S$
    }
  
    \SetKwProg{Fn}{Function}{:}{}
    \Fn{\FRemove{$S$, $\mathcal{G}$}}{
        derive the communication timeline with compression strategy $S$ and detect the communication bubbles;
        remove uncompressed tensors from $\mathcal{G}$ communicated before bubbles \;
    }
    
    \SetKwProg{Fn}{Function}{:}{}
    \Fn{\FSum{$S$, $idx$}}{
        \textsf{candidates} = [$S$] \;
        \ForEach{$c_i \in \mathcal{C}_{gpu}$} {
            $S_i = S.copy()$ \;
            $S_i[idx] = c_i$ \;
            $\textsf{candidates.add}(S_i)$ \;
        }
        \tcp{$F(S)$ is the iteration time with $S$}
        \KwRet $\arg \min \{ F(S_j) ~ | ~ S_j \in \textsf{candidates} \} $  \;
    }
\end{algorithm}

Algorithm~\ref{algo:gpu} shows {\name}'s GPU compression decision algorithm to determine the compression option of each tensor in a DNN model.
It first sorts and groups tensors with Lines 2-3 (Property \#2) and then rules out uncompressed tensors communicated before bubbles with \texttt{Remove()} (Property \#1).
Given a tensor $T_{idx}$, \texttt{GetBestOption()} enumerates all possible GPU compression options for this tensor and keeps the options of other tensors unchanged (Line 16-20).
Then there are $|\mathcal{C}_{gpu}| + 1$ strategy candidates (one of them is no compression). 
{\name} can derive the iteration time of each candidate with the empirical models introduced in Section~\ref{sec:interaction}.
Line 21 accounts for the interactions among tensors and selects the best candidate with the minimum iteration time (Property \#3).
After determining the compression option of one tensor, {\name} checks if new bubbles appear and rules out uncompressed tensors communicated before them again in Line 8 (Property \#1).

\subsubsection{{\name}'s CPU offloading}

{\name} offloads compression from GPUs to CPUs to further improve the training throughput of DDL after Algorithm~\ref{algo:gpu}.
Tensors with no compression are ruled out for CPU offloading and the set of the left tensors is denoted as $\mathcal{T}_{gpu}$, which can have hundreds of tensors.
The time complexity with brute force for CPU offloading is $O(2^{|\mathcal{T}_{gpu}|})$.
Tensors in $\mathcal{T}_{gpu}$ can have the same compression option, i.e., they take the same compression choice and communication schemes.
{\name} takes a greedy algorithm to find a provably optimal compression strategy for CPU offloading based on an interesting observation.



\vspace{-0.05in}
\begin{lemm}
    \label{lemma:cpu}
    Suppose $G$ is a set of tensors with the same size and same compression option from $\mathcal{T}_{gpu}$.
    Suppose also $q$ tensors in $G$ must be offloaded to CPUs for compression. The best solution is to offload the $q$ tensors farthest from the output layer.
\end{lemm}
\vspace{-0.05in}

The intuition of Lemma~\ref{lemma:cpu} is that offloading tensors to CPUs earlier can overlap more CPU compression with communication and tensor computation, and thus reduce the CPU compression overheads.
Therefore, if tensors in $\mathcal{T}_{gpu}$ can be grouped like $G$ in Lemma~\ref{lemma:cpu},
there is no need to evaluate all possible combinations because Lemma~\ref{lemma:cpu} restricts the choices of tensors for CPU offloading in each group.

\noindent \textbf{Algorithm 2.} {\name} first groups $\mathcal{T}_{gpu}$ to have $\mathcal{G}^{gpu} = \{G_1^{gpu}, G_2^{gpu}, \cdots, G_d^{gpu} \}$, where $G_i^{gpu}$ is a set of tensors with the same size and the same compression option. 
The tensors in $G_i^{gpu}$ are sorted in the descending order of their distances to the output layer.
Denote $U = \{u_1, u_2, \cdots, u_d\}$, where $u_i$ is the number of tensors in $G_i^{gpu}$ for CPU offloading and $0 \le u_i \le |G_i^{gpu}|$.
$\mathcal{U}$ is the set of all possible $U$.
For each $U \in \mathcal{U}$, {\name} considers a compression strategy that offloads the compression of the first $u_i$ tensors in $G_i^{gpu}$ to CPUs, and derives its iteration time.
It traverses $\mathcal{U}$ to search for the best $U$ with the minimum iteration time.

    
 


\vspace{-0.05in}
\begin{theore}
    \label{the:cpu}
    Algorithm 2 can find the best CPU offloading solution in $O(\prod (|G_i^{gpu}|+1))$ given $\mathcal{T}_{gpu}$.
\end{theore}
\vspace{-0.05in}



\vspace{-0.1in}

\section{Evaluation}

\subsection{Experimental Setup}
\label{exp_setup}
\noindent \textbf{Testbeds.} 
Two testbeds are used: 1) 8 GPU machines with NVLink and a 100Gbps network with TCP/IP, and 2) 8 PCIe-only GPU machines with a 25Gbps network.
Each machine has 8 NVIDIA Tesla V100 GPUs (32 GB GPU memory) and 2 CPUs/48 cores (Intel Xeon 8260 at 2.40GHz). 
Each machine runs Debian 10 and the software environment includes CUDA-11.0, PyTorch-1.8.0, BytePS-0.2.5, and NCCL-2.7.8.

\noindent \textbf{Workloads.}
We use six popular real-world DNN models including three computer vision models (VGG16, ResNet101 and UGATIT) and three NLP models (BERT-base, GPT2, and LSTM) by following the literature~\cite{byteps, omnireduce, SwitchML}. 
We set the batch sizes of these models by also following the literature~\cite{koliousis2019crossbow, merity2017regularizing, HiPress, omnireduce, SwitchML}. Specifically, the per-GPU batch size is kept constant as the number of GPUs increases, and the batch sizes are modest because large batch sizes are known to cause convergence problems~\cite{rethinkingCV, SwitchML}.
The details of the models, datasets, and batch sizes are shown in Table~\ref{table:models}.

\noindent \textbf{Compression algorithms.}
We use three representative compression algorithms:
Randomk~\cite{stich2018sparsified} and DGC~\cite{dgc} for sparsification (1\% compression rate), and EFSignSGD~\cite{efsignsgd} for quantization.
Error-feedback~\cite{efsignsgd, dgc} is applied on both GPU and CPU compression to preserve the model accuracy.

\noindent \textbf{Baselines.}
We use BytePS~\cite{byteps} as the training baseline without GC (FP32).
We use HiPress~\cite{HiPress} and HiTopKComm~\cite{HiTopKComm} as the two baselines with GPU compression, and BytePS-Compress~\cite{zhong2021compressed} as the baseline with CPU compression.

\noindent \textbf{Performance metrics.} 
We use trained images per second as the metric for computer vision  models and tokens per second for NLP models.
We measure the computational time of {\name} and training accuracy of DNN models.
We also provide the upper bound on the training throughput of compression-enabled DDL (Upper Bound).
This is obtained by assuming GC has no compression time and has no impact on tensor computation.


\noindent \textbf{Implementation.} 
We implement a GPU compression library shared by HiPress, HiTopKComm, and {\name} as well as a CPU compression library shared by BytePS-Compress and {\name}.
We also implement a communication library to support different communication schemes in both intra- and inter-machine communications shared by all baselines and {\name}.
These libraries consist of 5.1K and 3.0K lines of code in C++ and Python.
{\name}'s decision algorithm is implemented with 1.1K lines of code in Python.


\vspace{-0.15in}

\subsection{End-to-End Experiments}

\begin{figure*}[ht!]
    \centering
	\begin{subfigure}[t]{0.31\linewidth}
	\includegraphics[width=\linewidth]{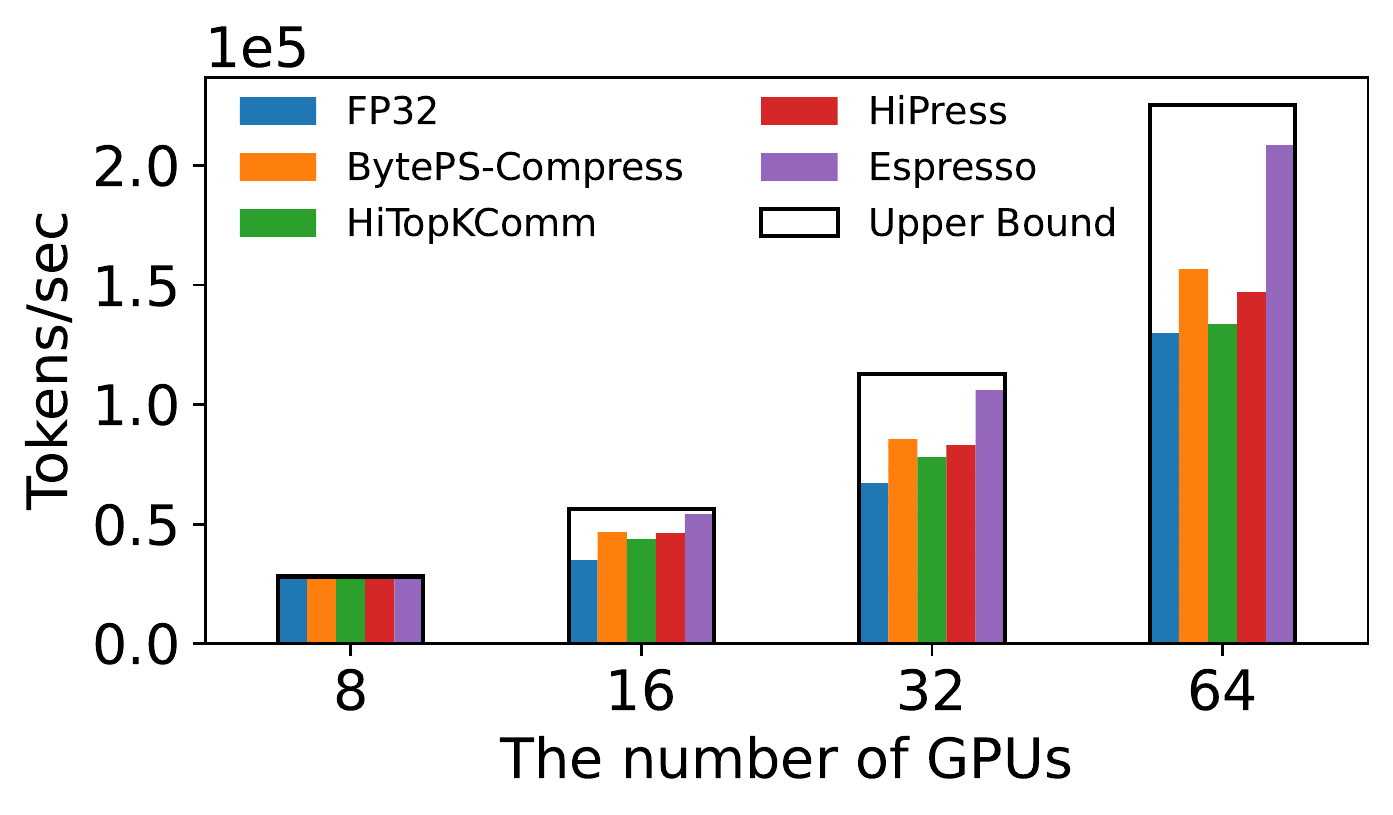}
	\vskip -0.1in
	\caption{BERT-base+Randomk}
	\label{fig:nvlink_bert_randomk}
    \end{subfigure}  
    \begin{subfigure}[t]{0.325\linewidth}
	\includegraphics[width=\linewidth]{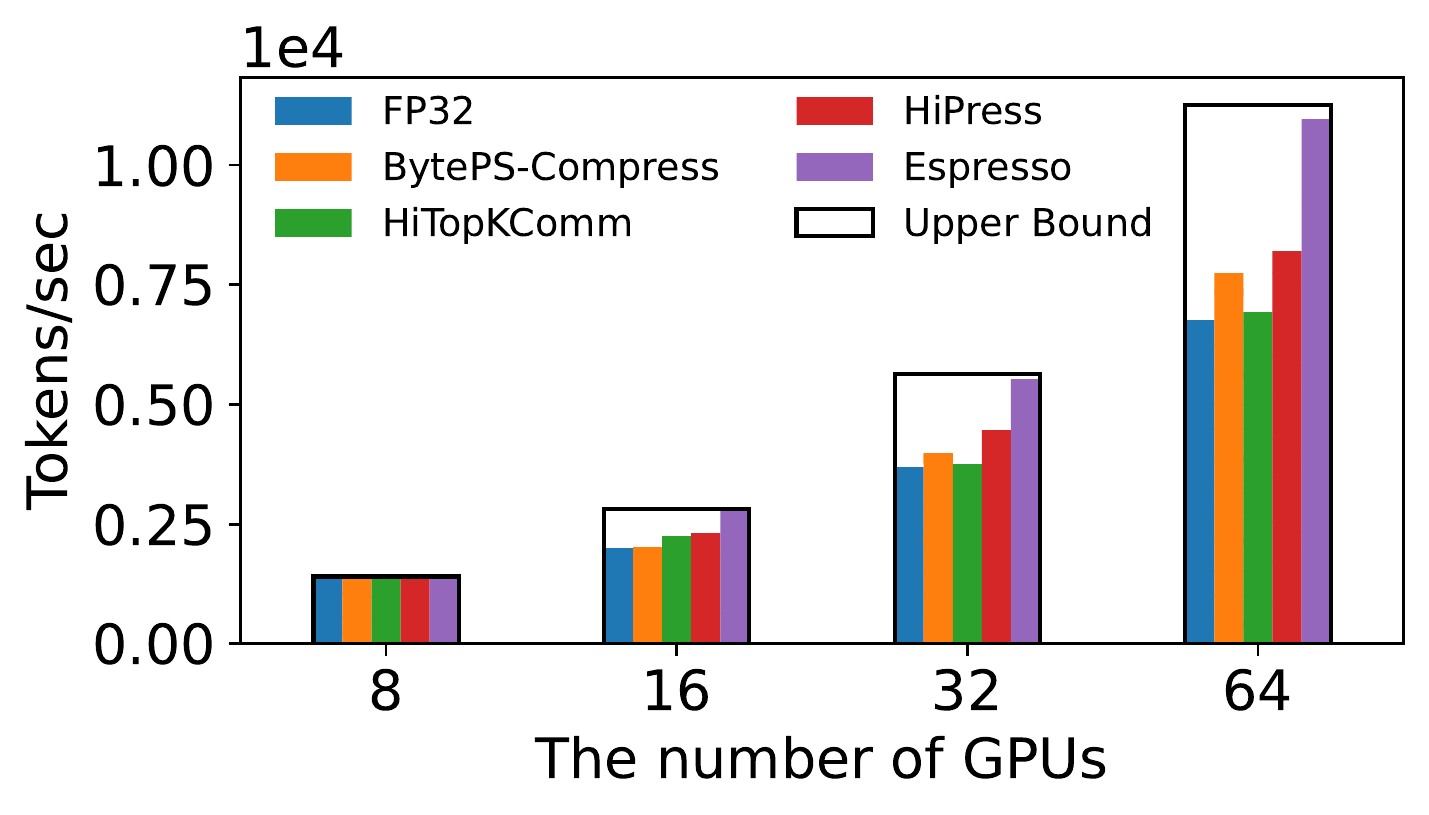}
	\vskip -0.1in
	\caption{GPT2+EFSignSGD}
	\label{fig:nvlink_gpt2_efsignsgd}
    \end{subfigure} 
    \begin{subfigure}[t]{0.325\linewidth}
	\includegraphics[width=\linewidth]{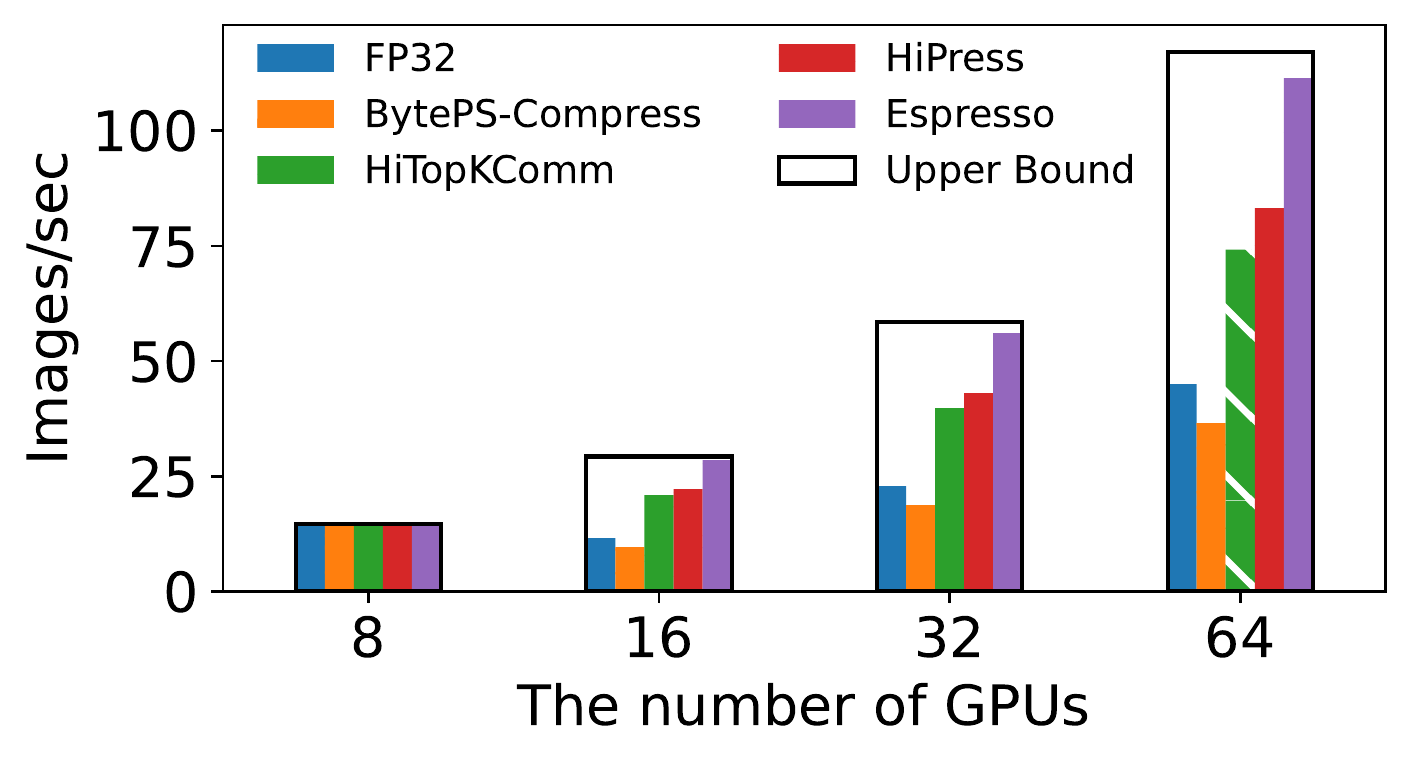}
	\vskip -0.1in
	\caption{UGATIT+DGC}
	\label{fig:nvlink_ugatit_dgc}
    \end{subfigure}    
    \vskip -0.15in
    \caption{Throughput of different DNN models with NVLink-based GPU machines and 100Gbps cross-machine Ethernet.}
    \label{fig:v100_nvlink}
    \vskip -0.2in
\end{figure*}

\begin{figure*}[ht!]
    \centering
     \begin{subfigure}[t]{0.325\linewidth}
	\includegraphics[width=\linewidth]{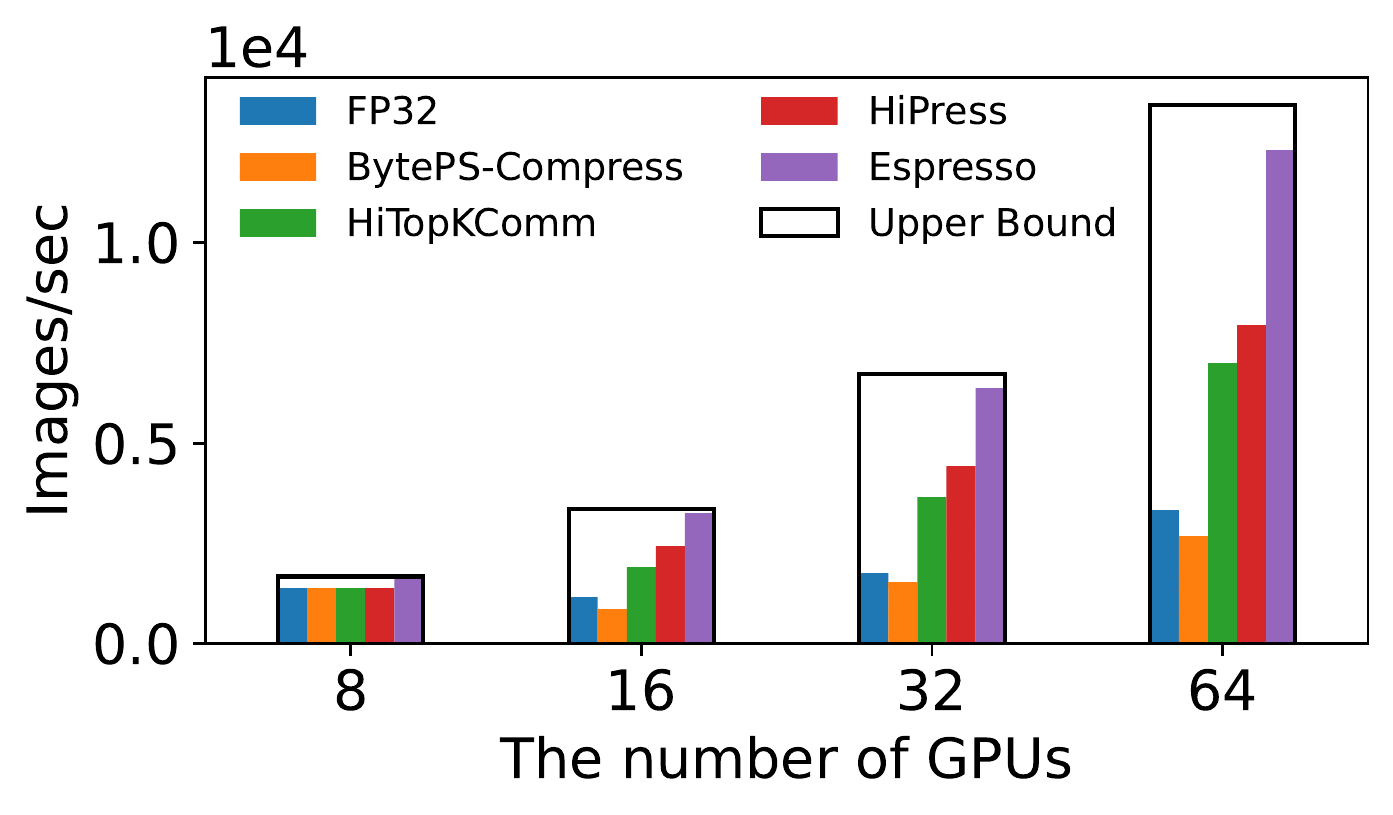}
	\vskip -0.1in
	\caption{VGG16+Randomk}
	\label{fig:pcie_vgg_randomk}
    \end{subfigure} 
	\begin{subfigure}[t]{0.31\linewidth}
	\includegraphics[width=\linewidth]{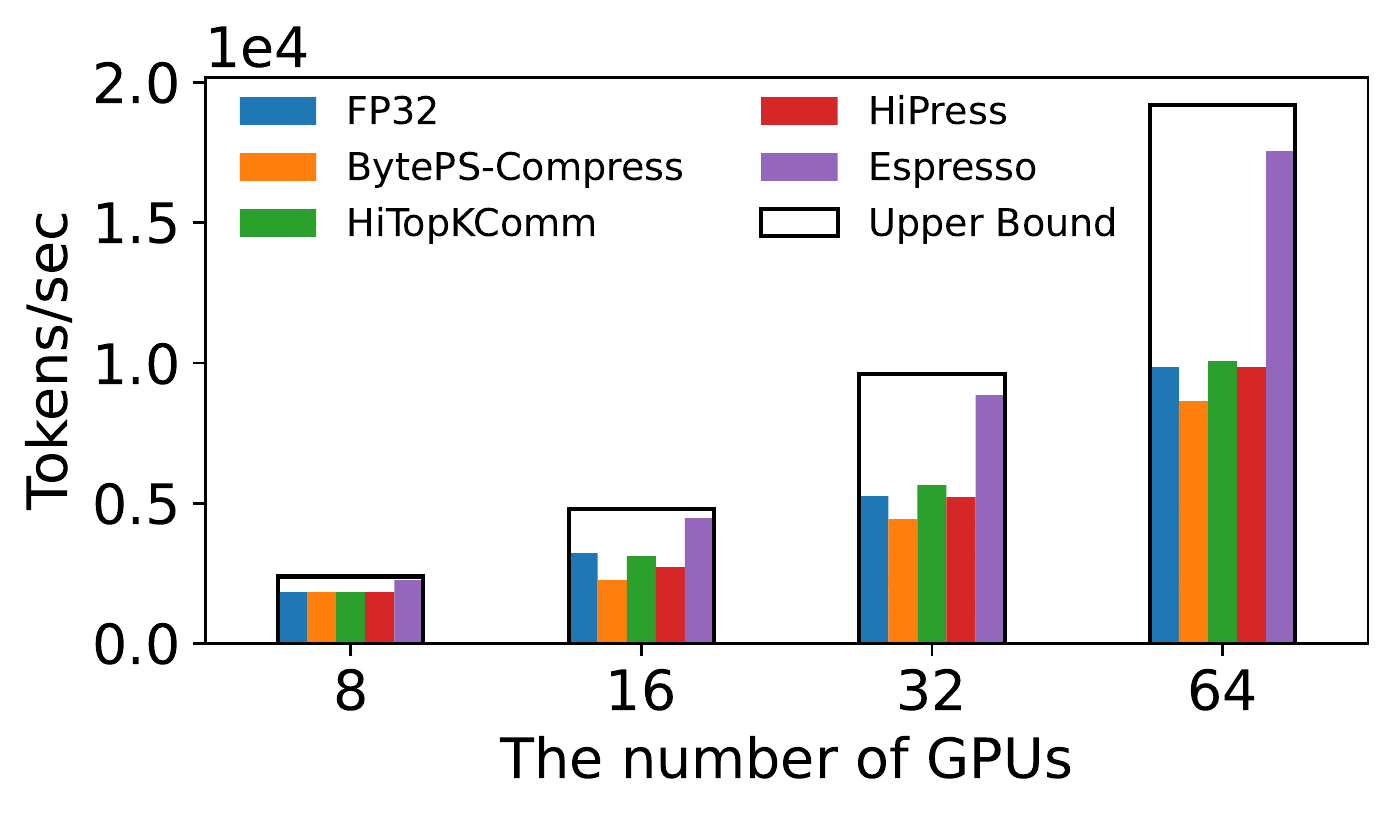}
	\vskip -0.1in
	\caption{LSTM+EFSignSGD}
	\label{fig:pcie_LSMT_efsignsgd}
    \end{subfigure}  
    \begin{subfigure}[t]{0.325\linewidth}
	\includegraphics[width=\linewidth]{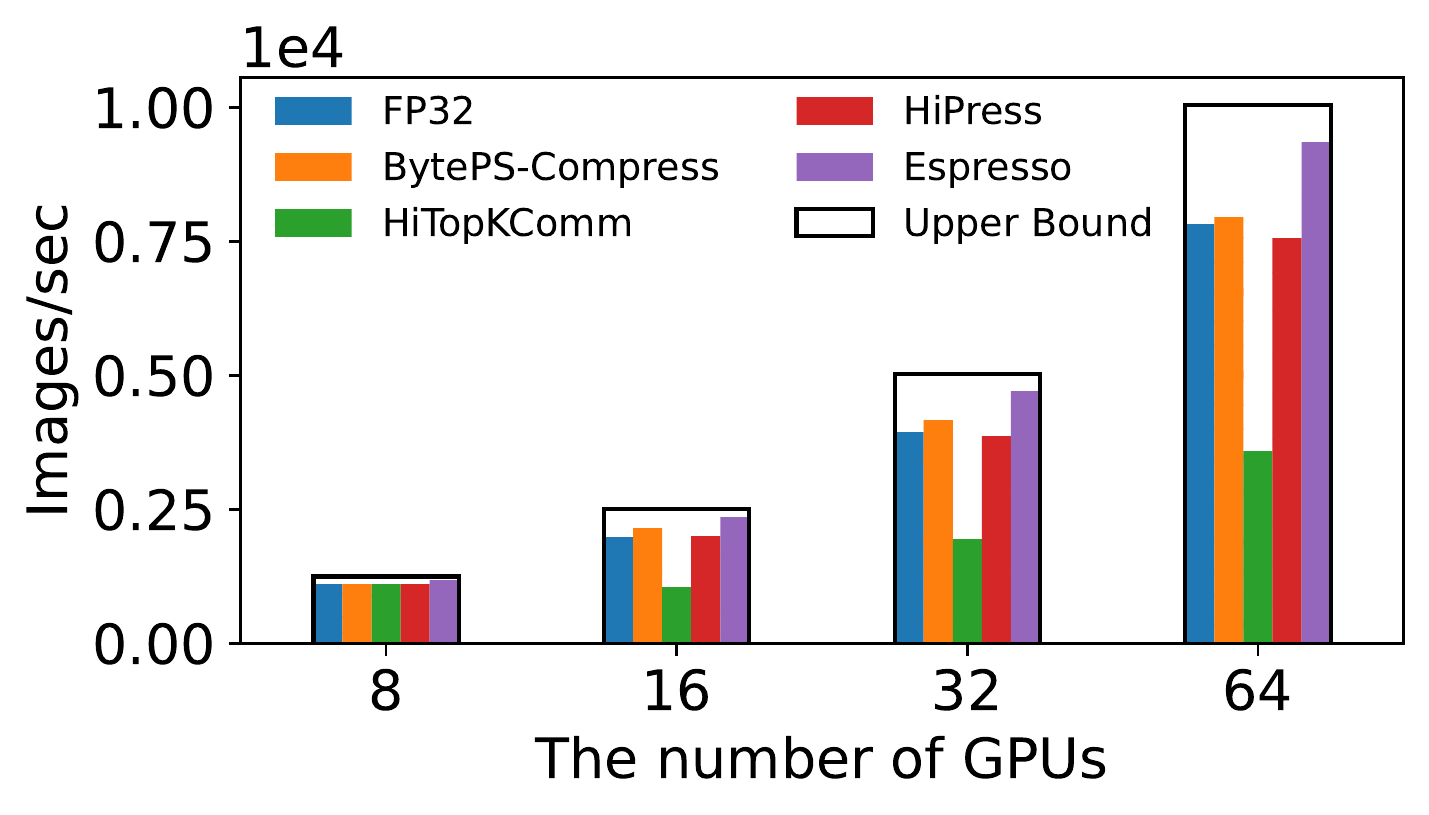}
	\vskip -0.1in
	\caption{ResNet101+DGC}
	\label{fig:pcie_resnet101_dgc}
    \end{subfigure}  
    \vskip -0.15in
    \caption{Throughput of different DNN models with PCIe-only GPU machines and 25Gbps cross-machine Ethernet.}
    \label{fig:v100_pcie}
\end{figure*}








\begin{table}[t!]
\scriptsize
\centering
    \begin{tabular}{l|lll}
    \hline
    Model    & Dataset     & Batch size & Model size  \\ \hline \hline
    VGG16~\cite{vgg}  & ImageNet~\cite{imagenet} & 32 images       & 528 MB   \\ \hline
    ResNet101~\cite{ResNet-50} & ImageNet~\cite{imagenet} & 32 images        & 170 MB     \\ \hline
    UGATIT~\cite{ugatit} & selfie2anime~\cite{selfie2anime} & 2 images & 2559 MB    \\ \hline
    BERT-base~\cite{bert}  & SQuAD~\cite{squad}       & 1024 tokens          & 420 MB  \\ \hline
    GPT2~\cite{gpt2} & WikiText-2~\cite{wikitext}  & 80 tokens          & 475 MB     \\ \hline
    LSTM~\cite{LSTM} & WikiText-2~\cite{wikitext}       & 80 tokens & 328 MB    \\ \hline
    \end{tabular}
    \caption {Characteristics of the benchmark DNN models.}
    \vskip -0.2in
    \label{table:models}
\end{table}

\begin{table}[t!]
\scriptsize
\centering
    \begin{tabular}{l|rrrrrr}
    \hline
                     &  VGG16     & ResNet101 & UGATIT & BERT-base & GPT2 & LSTM \\ \hline \hline
    \# of Tensors    &  32        & 314       & 148    & 207 & 148        & 10   \\ \hline
    {\name}          &  7ms       & 179ms     & 84ms   & 125ms      & 99ms    & 1ms   \\ 
    Brute force      &  >~24h     & >~24h     &  >~24h & >~24h  & >~24h       & >~24h   \\ \hline
    \end{tabular}
    \caption {The time to select compression strategies. \# of tensors is the number of tensors in DNN models.}
    \label{table:compute_strategy}
    \vskip -0.2in
\end{table}

\begin{table}[t!]
\scriptsize
\centering
    \begin{tabular}{l|rrrrrr}
    \hline
                     &  VGG16     & ResNet101 & UGATIT & BERT-base & GPT2  & LSTM \\ \hline \hline
    \# of Tensors    &  11        & 42        & 32     &   54      & 34      & 5   \\ \hline
    {\name}          &  1ms      &  14ms      & 1ms    &   2ms     & 1ms     & 1ms   \\ 
    Brute force      &  1ms      & >~24h      & 1.9h   & >~24h     & 7.6h    & 1ms   \\ \hline
    \end{tabular}
    \caption {The time to find best CPU offloading solutions. \# of tensors is the number of tensors for offloading.}
    \label{table:cpu_offloading}
    \vskip -0.2in
\end{table}

\subsubsection{DDL with NVLink-based GPU machines.}
Figure~\ref{fig:v100_nvlink} shows the training throughput of three DNN models with {\name} and baselines.
The performance bottleneck is inter-machine communication. 

As shown in Figure~\ref{fig:nvlink_bert_randomk}, the compression baselines bring very limited speedups over FP32 for BERT-base.
For example, HiTopKComm and HiPress only outperform FP32 by up to 4\% and 13\%, respectively.
It is because there are a large number of tensors in BERT-base, while none of the baselines consider the interactions among tensors.
Their compression strategies lead to costly compression overheads.
{\name} significantly improves the performance over all baselines.
For example, with 64 GPUs, it outperforms BytePS-Compress, HiTopKComm, and HiPress by 31\%, 54\%, and 40\%, respectively.
For GPT2, it outperforms BytePS-Compress and HiPress by 42\% and 33\% with 64 GPUs, as shown in Figure~\ref{fig:nvlink_gpt2_efsignsgd}.

UGATIT is very communication-intensive because of its large model size.
When the number of GPUs is 64, the performance improvement with HiPress and HiTopKComm is 86\% and 66\%, respectively, as shown in Figure~\ref{fig:nvlink_ugatit_dgc}.
BytePS-Compress even harms the performance by 18\% due to the costly computational overhead for CPU compression.
{\name} leverages both GPUs and CPUs for compression.
It outperforms FP32, BytePS-Compress, HiTopKComm, and HiPress by 149\%, 205\%, 50\%, and 35\%, respectively.
One important observation is that the improvements of {\name} become larger from 8 GPUs to 64 GPUs.
This implies that when DDL scales out, the computational overhead caused by compression also increases, and {\name} becomes more beneficial. 

\begin{figure}[t!]
    \centering
    \begin{subfigure}[t]{0.49\linewidth}
	\includegraphics[width=\linewidth]{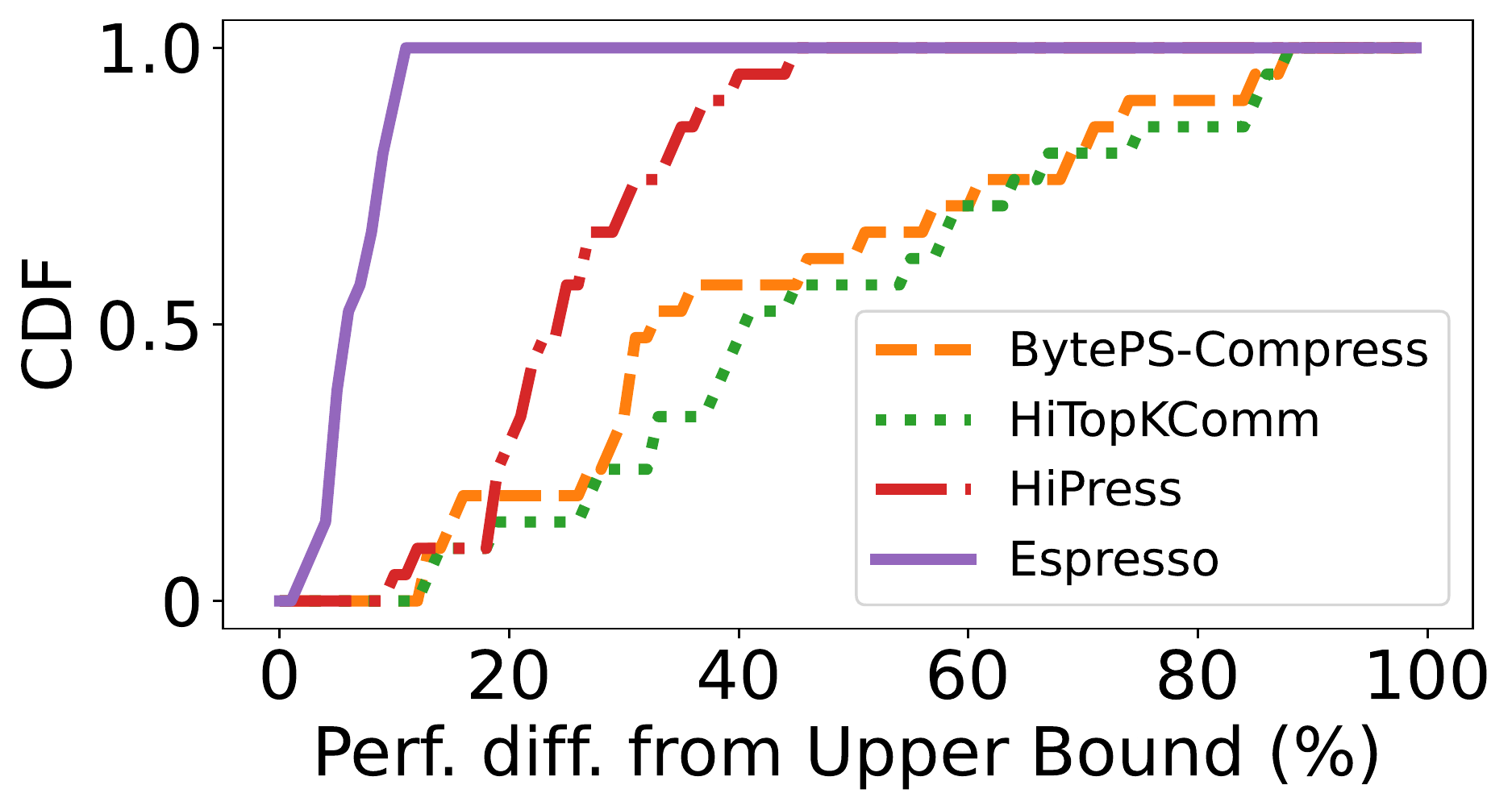}
	\vskip -0.05in
	\caption{NVLink-based machines}
	\label{fig:nvlink_upper_bound}
    \end{subfigure}  
    \begin{subfigure}[t]{0.49\linewidth}
	\includegraphics[width=\linewidth]{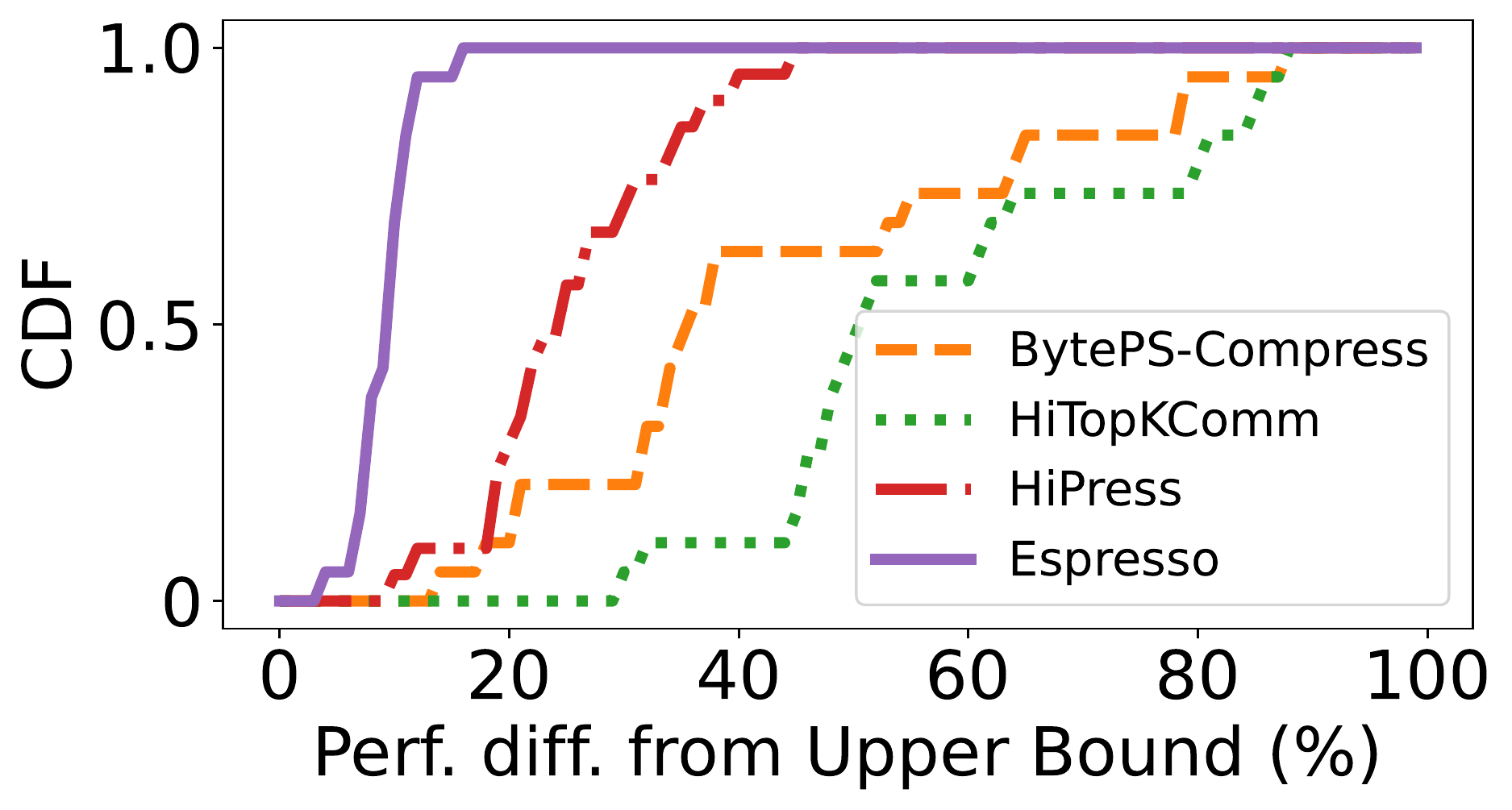}
	\vskip -0.05in
	\caption{PCIe-only machines}
	\label{fig:pcie_upper_bound}
    \end{subfigure}    
    \vskip -0.15in
    \caption{The performance differences between compression frameworks and Upper Bound with 64 GPUs.}
    \label{fig:upper_bound}
\end{figure}


\subsubsection{Computational time of {\name}.}

Table~\ref{table:compute_strategy} lists the computational time of {\name} to select compression strategies for the training of different DNN models with 8 NVLink-based GPU machines (the results are similar with PCIe-only GPU machines).
The time increases with the number of tensors in DNN models, but even for ResNet101 with 314 tensors, the computational time is still within one iteration time.
In contrast, brute force takes a very long time because it has to traverse all the possibilities. 
Even though LSTM only has 10 tensors, the searching time is still unacceptable.

Table~\ref{table:cpu_offloading} shows the computational time of {\name} to find the best CPU offloading solution.
After {\name}'s GPU compression decision algorithm, the number of tensors for CPU offloading has been significantly reduced.
Brute force can quickly find the best solution for VGG16 and LSTM, but it takes a long time for other models.
{\name} can still quickly find the best solution.
For example, there are 54 tensors in BERT-base for CPU offloading, but they only have a few different tensor sizes, as shown in Figure~\ref{fig:tensor_size}.
{\name} only needs to consider a few thousand choices to find the best CPU offloading.

\begin{figure*}[t!]
    \centering
    \begin{subfigure}[t]{0.24\linewidth}
	\includegraphics[width=\linewidth]{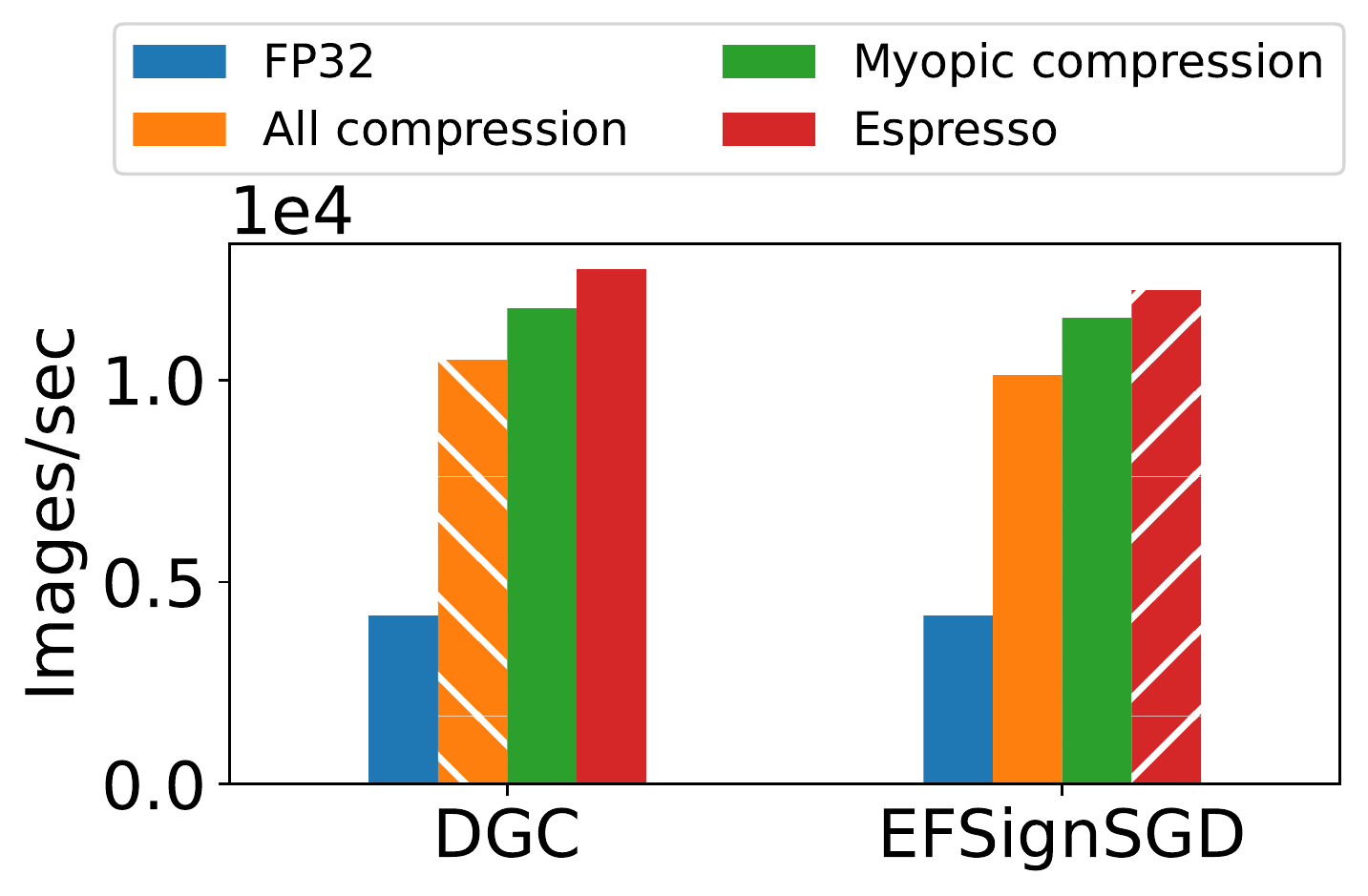}
	\vskip -0.1in
	\caption{Restrict Dim. 1}
    \end{subfigure}    
	\begin{subfigure}[t]{0.24\linewidth}
	\includegraphics[width=\linewidth]{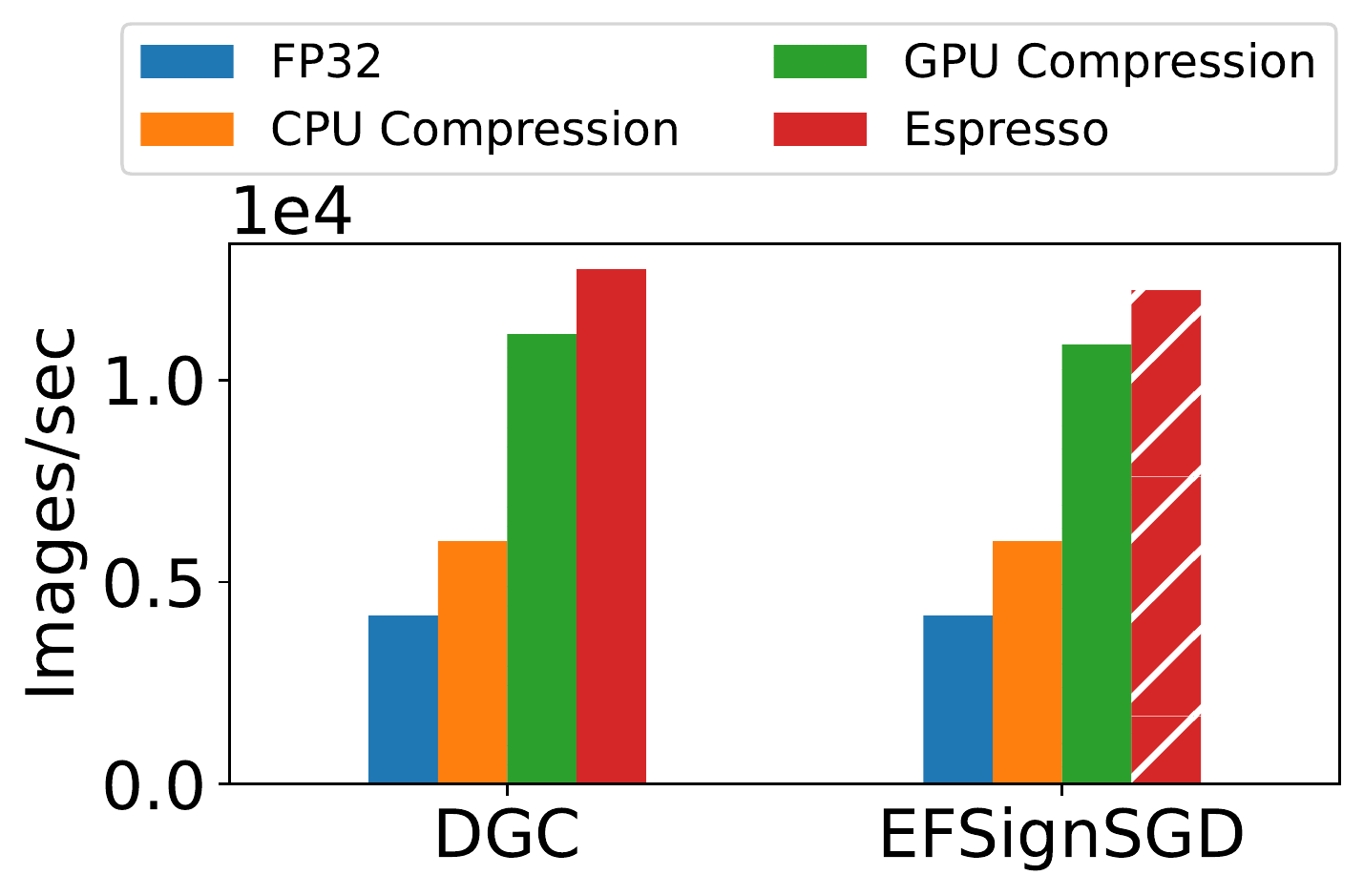}
	\vskip -0.1in
	\caption{Restrict Dim. 2}
    \end{subfigure}  
    \begin{subfigure}[t]{0.24\linewidth}
	\includegraphics[width=\linewidth]{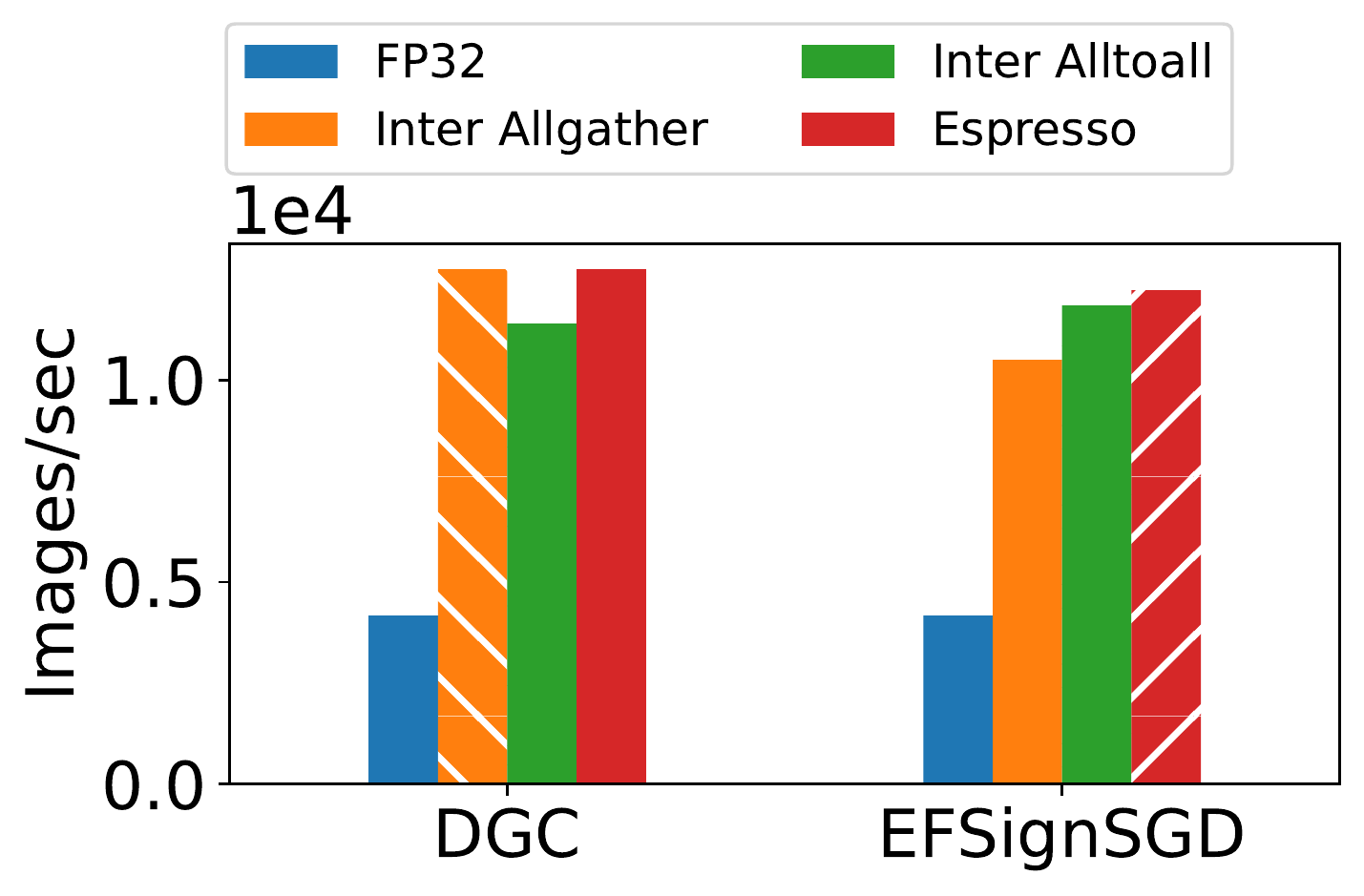}
	\vskip -0.1in
	\caption{Restrict Dim. 3}
    \end{subfigure}  
    \begin{subfigure}[t]{0.24\linewidth}
	\includegraphics[width=\linewidth]{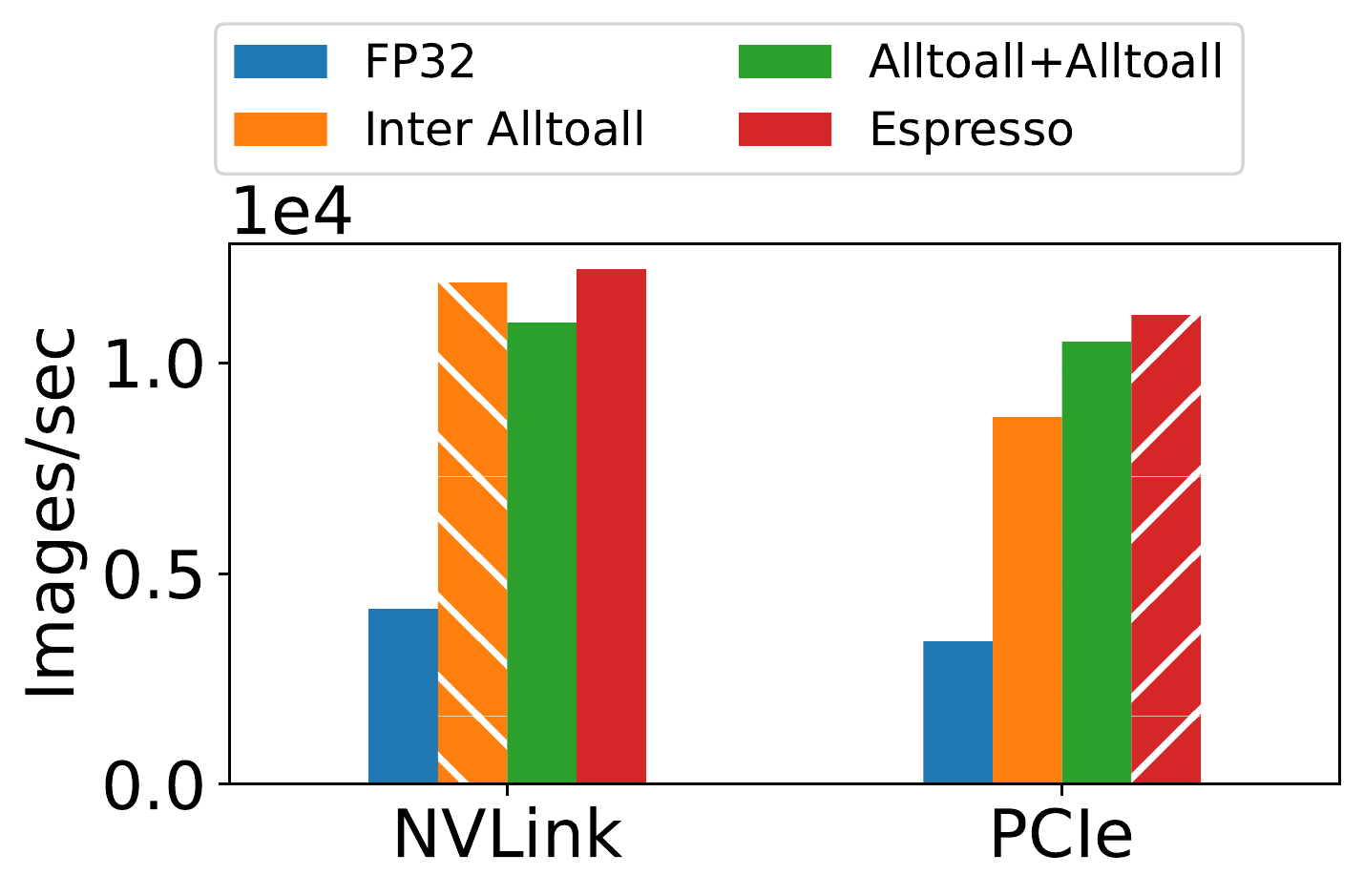}
	\vskip -0.1in
	\caption{Restrict Dim. 4}
    \end{subfigure}  
    \vskip -0.15in
    \caption{Considering all 4 dimensions is always better than considering only 3 dimensions.}
     \vskip -0.15in
     \label{fig:cripple}
\end{figure*}

\vspace{-0.15in}

\subsubsection{DDL with PCIe-only GPU machines.}
The performance bottlenecks could be both inter- and intra-machine communications in this setup.
Figure~\ref{fig:pcie_LSMT_efsignsgd} shows that the three compression baselines bring almost no improvement for LSTM model with GC.
For example, HiPress only outperforms FP32 by up to 2\%, and BytePS-Compress even harms the performance by 12\% with 64 GPUs.
It is because they only compress tensors to reduce inter-machine communication and cannot effectively alleviate the intra-machine communication bottleneck.
Moreover, they also incur costly compression overhead.
{\name} compresses tensors to reduce both inter- and intra-machine communications when necessary and always has the best performance across all cases.
For example, with 64 GPUs, it outperforms BytePS-Compress, HiTopKComm, and HiPress by 101\%, 73\%, 77\%, respectively.
For VGG16 model with 64 GPUs, the speedups of {\name} over FP32, BytePS-Compress, and HiPress are 269\%, 357\%, 55\%, respectively.

We observe that ResNet101 is not communication-intensive and it achieves the scaling factor of 0.70 with FP32.
Figure~\ref{fig:pcie_resnet101_dgc} shows applying GC to ResNet101 with the compression baselines can harm its performance. 
HiTopKComm reduces its training throughput by up to 54\% because it compresses all the tensors and leads to exorbitant compression overhead.
HiPress also has high over-compression penalties and it degrades the performance by 4\% with 64 GPUs.
In contrast, {\name} still outperforms FP32, BytePS-Compress, and HiPress by up to 20\%, 18\%, and 24\%, respectively.


\vspace{-0.12in}

\sloppy{\subsubsection{{\name}'s compression strategies are near-optimal}
We have performed experiments for all combinations of GC} algorithms (i.e. Randomk, DGC, EFSignSGD), DNN models (i.e. VGG16, ResNet101, UGATIT, BERT-base, GPT2, LSTM), varying the number of GPUs from 8 to 64, over both NVLink and PCIe, across all schemes (i.e. FP32, HiPress, BytePS-Compress, HiTopKComm, Espresso). 
However, due to space limitations, we present a summary of all the results for the 64-GPU scenario; the raw results will be included in a technical report in the future. Specifically, we present the cumulative distribution of the performance differences of each scheme from the Upper Bound.
Figure~\ref{fig:nvlink_upper_bound} displays the distributions of performance differences for all the training with NVLink-based machines and 64 GPUs.
The performance differences between {\name} and Upper Bound is always less than 10\%. 
To call out a few specific data points, the performance differences for the training of GPT2 with EFSignSGD, UGATIT with DGC, and BERT-base with Randomk are only 3\%, 5\%, and 7\%, respectively. 
Note that the differences between {\name}'s compression strategy and the optimal strategy can be even smaller because Upper Bound is by definition higher than the training throughput of the optimal strategy.
Figure~\ref{fig:pcie_upper_bound} shows the distributions for all the training with PCIe-only machines and 64 GPUs and Espresso similarly out-performs other baselines.

\vspace{-0.15in}

\subsection{Importance of the Entire Search Space}


To evaluate the importance of considering all four dimensions,
we cripple one of the dimensions and then select the compression strategy with the remaining three dimensions.
We cripple Dimension 1 with two restricted mechanisms: \textbf{All compression:} It compresses all tensors. \textbf{Myopic compression:}
It does not consider interactions among tensors when applying GC to tensors.
We cripple Dimension 2 with two restricted mechanisms: \textbf{GPU compression:}
It only compresses tensors with GPUs. \textbf{CPU compression:}
It only compresses tensors with CPUs.
\sloppy{We cripple Dimension 3 with two restricted mechanisms: \textbf{Inter Allgather:}
It compresses tensors for inter-machine communication and uses Allgather for compressed tensors. \textbf{Inter Alltoall:}
It compresses tensors for inter-machine communication. The communication scheme is Alltoall/Allgather.}
We cripple Dimension 4 with \textbf{Inter Alltoall} and another restricted mechanism \textbf{Alltoall+Alltoall:}
It first compresses tensors for the first intra-machine communication and the communication scheme is Alltoall.
It then decompresses and compresses tensors again for inter-machine communication.
It uses Alltoall/Allgather for inter-machine communication and Allgather for the second intra-machine communication.

Figure~\ref{fig:cripple} shows the scaling factors of VGG16 with 64 GPUs.
NVLink-based GPU machines are used in (a), (b), and (c), and EFSignSGD is used in (d).
The compression strategies determined by {\name} always outperforms the compression strategies selected from the cripple search space.
Moreover, Figure~\ref{fig:cripple}(c) verifies that different types of GC algorithms need different communication schemes, and Figure~\ref{fig:cripple}(d) verifies that different intra- and inter-machine bandwidth need different compression choices.

\vspace{-0.1in}

\subsection{Convergence validation}


\begin{figure}[t!]
    \centering
    \begin{subfigure}[t]{0.53\linewidth}
	\includegraphics[width=\linewidth]{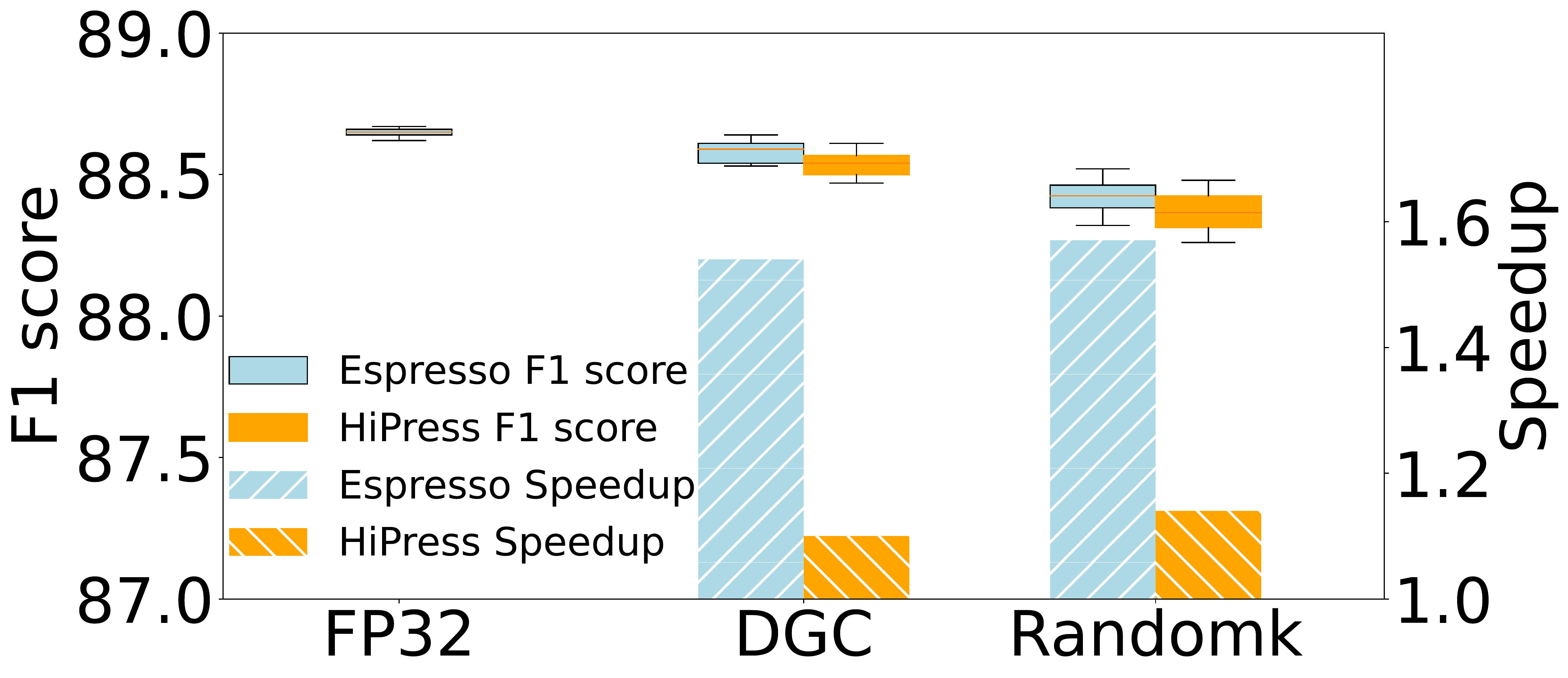}
	\vskip -0.05in
	\caption{BERT}
	\label{fig:bert_accuracy}
    \end{subfigure}  
    \begin{subfigure}[t]{0.44\linewidth}
	\includegraphics[width=\linewidth]{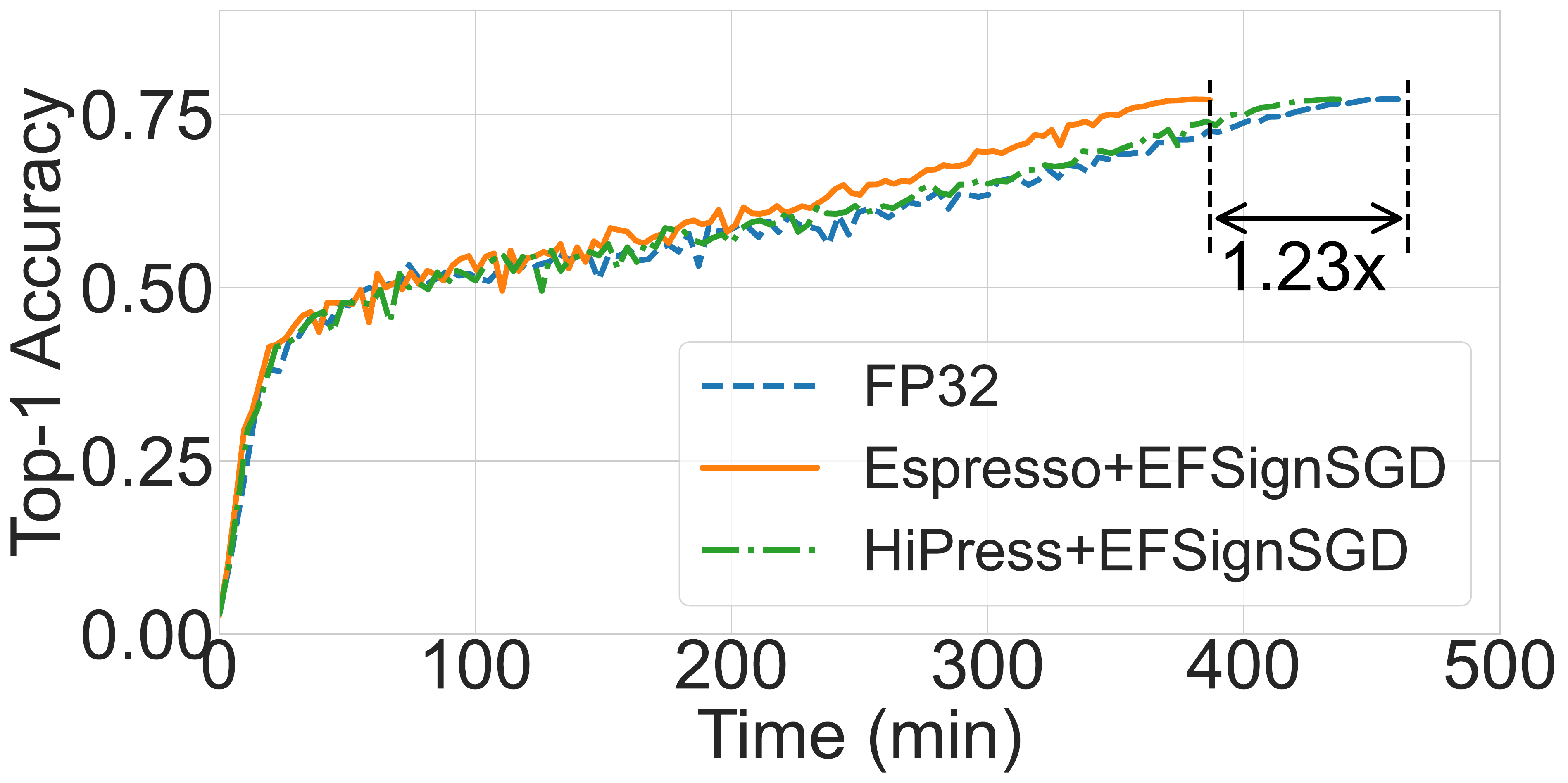}
	\vskip -0.05in
	\caption{ResNet101}
	\label{fig:resnet101_accuracy}
    \end{subfigure}    
    \vskip -0.15in
    \caption{Model accuracy of BERT-base (F1 score) and ResNet101 (Top-1 accuracy).}
    \label{fig:accuracy}
\end{figure}

It has been theoretically proven and empirically validated that GC can preserve the training accuracy and convergence~\cite{1bit, wu2018error, stich2018sparsified, dgc, jiang2018linear, omnireduce, HiPress}.
In this section, we reaffirm these conclusions and demonstrate that {\name} can preserve the training accuracy and convergence.

We conduct a test following the methodology in~\cite{omnireduce} to fine-tune BERT-base for the question answering task on SQuAD~\cite{squad} for two epochs and repeat the experiments ten times.
The number of GPUs is 64 on 8 NVLink-based GPU machines.
Figure~\ref{fig:bert_accuracy} shows that {\name} with DGC can achieve around 1.55$\times$ speedup over no compression (i.e. FP32) and it has almost the same F1 score as no compression.
We also train ResNet101 for 120 epochs on ImageNet~\cite{imagenet} from scratch and apply EFSignSGD to the model training. 
As shown in Figure~\ref{fig:resnet101_accuracy}, the speedup of {\name} over no compression (i.e. FP32) is 1.23$\times$.
The achieved Top-1 accuracy with {\name} is 77.10\%, which is very close to the no-compression accuracy of 77.18\%.






\vspace{-0.1in}

\section{Related Work}

GRACE~\cite{xu2021grace} quantitatively evaluates the impacts of GC algorithms and observes that GC can incur non-negligible compression overhead, but it does not study or address the challenges of applying GC to DDL.
Several frameworks are recently proposed to support compression-enabled DDL.
HiTopKComm~\cite{HiTopKComm} designs a new communication scheme for GC, but it compresses all tensors with GPUs and leads to prohibitive compression overhead.
HiPress~\cite{HiPress} proposes compression-aware synchronization to overlap compression with communication and a selective compression mechanism to decide
whether to compress a tensor, but it only uses GPUs for compression and ignores the interactions among tensors.
BytePS~\cite{zhong2021compressed} also supports GC, but it only uses CPUs for compression and ignores the interactions among tensors as well.
These frameworks only compress tensors for inter-machine communication.
In contrast, {\name} uses both GPUs and CPUs for compression, analyses interactions among tensors to make compression decisions, and address both intra- and inter-machine communication bottlenecks.
OmniReduce~\cite{omnireduce} introduces block gradient sparsification, which is a new type of GC algorithm, but {\name} focuses on how to efficiently apply GC to DDL.





Other than GC~\cite{aji2017sparse, dgc, adacomp, sparse2018gradient, wen2017terngrad, 1bit, efsignsgd, signsgd, dist-ef-sgd, QSGD}, there are other approaches that aim at improving the training throughput of DDL. 
SwitchML~\cite{SwitchML} and ATP~\cite{atp} exploit programmable switches for gradient aggregations.
Other communication schemes have been proposed to more efficiently aggregate gradients.
For example, BytePS~\cite{byteps} uses spare CPU and bandwidth resources in GPU clouds to optimize both intra- and inter-machine communications.
Blink~\cite{blink} generates optimal communication primitives for intra-machine communication with NVLink.
PLink~\cite{luo2020plink} designs a hierarchical aggregation scheme for DDL in public clouds, where the machine-to-machine bandwidth is non-uniform due to the hierarchical structure of data centers.
ByteScheduler~\cite{bytescheduler}, P3~\cite{p3}, and TicTac~\cite{tictac} 
schedule communications of tensors closer to the output layer with higher priority.
These approaches are compression-agnostic; since {\name} supports compression-enabled DDL, it can be integrated with most of them.

\vspace{-0.1in}
\section{Conclusion}
{\name} is a general framework to enable DDL to achieve near-optimal training speed with GC.
It holistically considers all the dimensions when making decisions for how to apply GC to DDL.
{\name} can express all the compression strategies and analyze the intricate interactions among tensors to quickly select a near-optimal compression strategy for any DDL training job.
It outperforms the state-of-the-art compression-enabled systems by up to 77\% across six popular DNN models and preserve model accuracy.

\bibliographystyle{ACM-Reference-Format}
\bibliography{reference}


\end{document}